\documentclass{article}
 
\usepackage{hyperref} 
\usepackage{float}    
\usepackage{amsmath, amsthm}  
\usepackage{amssymb}  
\usepackage{subcaption}   
\usepackage{xcolor}
\usepackage{booktabs} 
\usepackage[square, numbers]{natbib}
\usepackage{multirow}
\usepackage{algorithm, algcompatible}
\usepackage[margin=1.25in]{geometry}
\usepackage[final]{graphicx} 
\usepackage{xfrac} 
\usepackage{wrapfig}

\newtheorem*{theorem*}{Theorem}

\newcommand{\RR}{\mathbb{R}}
\newcommand{\EE}{\mathbb{E}}

\newcommand{\name}{\text{ARS}}

\DeclareMathOperator*{\diag}{diag}

\newcommand{\id}{\mathbf{I}}

\newcommand{\Acal}{\ensuremath{\mathcal{A}}}
\newcommand{\Ocal}{\ensuremath{\mathcal{O}}}
\newcommand{\Tcal}{\ensuremath{\mathcal{T}}}

\newcommand{\R}{\ensuremath{\mathbb{R}}}

\newcommand{\vone}{\textbf{V1}}
\newcommand{\vtwo}{\textbf{V2}}
\newcommand{\vtwot}{\textbf{V2-t}}
\newcommand{\vonet}{\textbf{V1-t}}

\newcommand{\statedim}{n}
\newcommand{\inputdim}{p}

\title{Simple random search provides a competitive approach\\ to reinforcement learning}

\author{Horia Mania \and Aurelia Guy \and Benjamin Recht \and \\
Department of Electrical Engineering and Computer Science\\
University of California, Berkeley}

\date{\today}

\begin{document} 

\maketitle

\begin{abstract}
A common belief in model-free reinforcement learning is that methods based on random search in the parameter space of policies exhibit significantly worse sample complexity than those that explore the space of actions. We dispel such beliefs by introducing a random search method for training static, linear policies for continuous control problems, matching state-of-the-art sample efficiency on the benchmark MuJoCo locomotion tasks. Our method also finds a nearly optimal controller for a challenging instance of the Linear Quadratic Regulator, a classical problem in control theory, when the dynamics are not known. 
Computationally, our random search algorithm is at least $15$ times more efficient than the fastest competing model-free methods on these benchmarks. 
We take advantage of this computational efficiency to evaluate the performance of our method over hundreds of random seeds and many different hyperparameter configurations for each benchmark task. Our simulations highlight a high variability in performance in these benchmark tasks, suggesting that commonly used estimations of sample efficiency do not adequately evaluate the performance of RL algorithms. 
\end{abstract}

\section{Introduction} 

Model-free reinforcement learning (RL) aims to offer off-the-shelf solutions for controlling dynamical systems 
without requiring models of the system dynamics. Such methods have successfully produced RL agents that surpass human players in video games and games such as Go \cite{mnih2015human, silver2016mastering}. Although these results are impressive, model-free methods have not yet been successfully deployed to control physical systems, outside of research demos. There are several factors prohibiting the adoption of model-free RL methods for controlling physical systems: the methods require too much data to achieve reasonable performance, the ever-increasing assortment of RL methods makes it difficult to choose what is the best method for a specific task, and many candidate algorithms are difficult to implement and deploy \cite{henderson2017deep}. 


Unfortunately, the current trend in RL research has put these impediments at odds with each other. In the quest to find methods that are \emph{sample efficient} (i.e. methods that need little data) the general trend has been to develop increasingly complicated methods. This increasing complexity has lead to a reproducibility crisis. Recent studies demonstrate that many RL methods are not robust to changes in hyperparameters, random seeds, or even different implementations of the same algorithm~\cite{henderson2017deep, islam2017reproducibility}. Algorithms with such fragilities cannot be integrated into mission critical control systems without significant simplification and robustification.

Furthermore, it is common practice to evaluate and compare new RL methods by applying them to video games or simulated continuous control problems and measure their performance over a small number of independent trials (i.e., fewer than ten random seeds) \cite{haarnoja2017reinforcement, haarnoja2018soft, heess2017emergence, lillicrap2015continuous,  mnih2016asynchronous, nagabandi2017neural, plappert2017parameter, rajeswaran2017towards, salimans2017evolution, schulman2015trust, schulman2015high, schulman2017proximal, silver2014deterministic, wang2016sample, wu2017scalable}. The most popular continuous control benchmarks are the MuJoCo locomotion tasks \cite{ 1606.01540, todorov2012mujoco}, with the Humanoid model being considered ``one of the most challenging 
continuous control problems solvable by state-of-the-art RL techniques \cite{salimans2017evolution}.'' In principle, one can use video games and simulated control problems for beta testing new ideas, but simple baselines should be established and thoroughly evaluated before moving towards more complex solutions.


To this end, we aim to determine \emph{the simplest} model-free RL method that can solve standard benchmarks. Recently, two different directions have been proposed for simplifying RL. 
\citet{salimans2017evolution} introduced a derivative-free policy optimization method, 
called Evolution Strategies (ES).
The authors showed that, for several RL tasks, their method can easily be parallelized to train policies faster than other methods.
While the method proposed by~\citet{salimans2017evolution} is simpler than previously proposed methods, it employs several complicated algorithmic elements, which we discuss in Section~\ref{sec:es}. 
As a second simplification to model-free RL, \citet{rajeswaran2017towards} have shown that linear policies can be trained via natural policy gradients to obtain competitive performance on the MuJoCo locomotion tasks, showing that complicated neural network policies are not needed to solve these continuous control problems. 
In this work, we combine ideas from the work of~\citet{salimans2017evolution} and~\citet{rajeswaran2017towards}
to obtain the simplest model-free RL method yet, a derivative-free optimization algorithm for training linear policies.
We demonstrate that a simple random search method
can match or exceed state-of-the-art sample efficiency on MuJoCo locomotion benchmarks. Moreover, our method is at least $15$ times more computationally efficient than ES, the fastest competing method. 
Our findings contradict the common belief that policy gradient techniques, which rely on exploration in the action space, are more sample efficient  than methods based on finite-differences \cite{peters2008reinforcement, plappert2017parameter}.
In more detail, our contributions are as follows:

\begin{itemize}   
  
\item 

In Section~\ref{sec:alg}, we present a classical, basic random search algorithm for solving derivative-free optimization
problems. For application to continuous control, we augment the basic random search method with three simple features.
First, we scale each update step by the standard deviation of the rewards collected for computing that update step. 
Second, we normalize the system's states by online estimates of their mean and standard deviation. Third, we discard from the computation of the update steps the directions that yield the least improvement of the reward. We refer to this method as \emph{Augmented Random Search}\footnote{Our implementation of ARS can be found at \url{https://github.com/modestyachts/ARS}.} (ARS).

\item In Section~\ref{sec:mujoco}, we evaluate the performance of ARS on the  
benchmark MuJoCo locomotion tasks.  Our method can learn static, linear policies that achieve high rewards on all MuJoCo tasks. That is, our control action is a linear map of the current states alone. No neural networks are used, and yet state-of-the-art performance is still uniformly achieved. For example, for the Humanoid model ARS finds linear policies which achieve average rewards of over $11500$, the highest reward reported in the literature.

To put ARS on equal footing with competing methods, we evaluated its required sample complexity to solve the MuJoCo locomotion tasks 
over three random seeds, uniformly sampled from an interval. 
We compare the measured performance of our method with results reported by \citet{haarnoja2018soft}, \citet{rajeswaran2017towards}, \citet{salimans2017evolution}, and \citet{schulman2017proximal}. 
ARS matches or exceeds state-of-the-art sample efficiency on the MuJoCo locomotion tasks.
 
 \item 
In Section~\ref{sec:computation} we report the time and computational resources required by ARS to train policies for the Humanoid-v1 task. We measure the time required to reach an average reward of $6000$ or more, and our results are reported over a hundred random seeds. On one machine with $48$ CPUs, ARS takes at most $13$ minutes on $25/100$ random seeds, and takes at most $21$ minutes on $50/100$ random seeds.  
Training policies for the  Humanoid-v1 task to reach the same reward threshold takes about a day on modern hardware with the popular Trust Region Policy Optimization (TRPO) method \cite{salimans2017evolution, schulman2015trust}, and takes around $10$ minutes with ES when parallelized over $1440$ CPUs \cite{salimans2017evolution}.  
Therefore, our method is at least $15$ times more computationally efficient than ES, the fastest competing method.

\item 
Since our method is more efficient than previous approaches, we are able to explore the variance of our method over many  random seeds. RL algorithms exhibit large training variances and hence evaluations over a small number of random seeds do not accurately capture their performance.  
\citet{henderson2017deep} and \citet{islam2017reproducibility} have already discussed the importance of measuring the performance of RL algorithms over many random seeds, and the sensitivity of RL methods to choices of hyperparameters.
For a more thorough evaluation of our method, we measured performance of ARS over a hundred random seeds and also evaluated its sensitivity to hyperparameter choices. 
Though ARS successfully trains policies for the MuJoCo locomotion tasks a large fraction of the time when hyperparameters and random seeds are varied, we note that it still exhibits a large variance, and that we still frequently find that learned policies do not uniformly yield high rewards.
 
\item 
In order to simplify and streamline the evaluation of RL for continuous control, we argue that it is important to add more baselines that are extensible and reproducible. In Section~\ref{sec:lqr} we argue for  using the Linear Quadratic Regulator (LQR) with unknown dynamics as such a benchmark. We evaluate the performance of ARS, over a hundred random seeds, on a difficult instance of this problem.
Although not as sample efficient as model-based methods, ARS finds nearly optimal solutions for the LQR instance considered.

\end{itemize}

\subsection{Related Work}

With the recent adoption of standard benchmark suites, a large body of recent research has applied RL methods for continuous control inside of simulation environments. \citet{levine2013guided} were among the first to use MuJoCo as a testbed for learning based control, and were able to achieve walking in complex simulators without special purpose techniques.  Since then, this simulation engine has been used by a variety of different researchers in different contexts to compare RL techniques. We list many of these approaches here, highlighting that the benefits and the comparisons of the approaches listed below were assessed over a small set of random seeds, often using unclear methods for hyperarameter selection. \citet{henderson2017deep} and \citet{islam2017reproducibility} 
 pointed out that such methodology does not accurately capture the performance of RL methods, which are sensitive to both the choice of random seed
and the choice of hyperarameters. 

\citet{mnih2016asynchronous} showed that actor-critic methods, 
popular for variance reduction in policy gradient algorithms,   
can be asynchronously parallelized for fast training of policies
for Atari video games and MuJoCo models. Previously, \citet{schulman2015high} introduced the Generalized Advantage Estimation (GAE) method for estimating advantages, offering  
variance reduction with less bias than previous methods.

The popular Trust Region Policy Optimization (TRPO) algorithm is related to the natural gradient method. 
TRPO, introduced by \citet{schulman2015trust}, maximizes at each iteration an approximate 
average reward objective regularized by a KL-divergence penalty. 
As a more scalable 
trust region method, \citet{wu2017scalable} proposed an actor critic method which 
uses Kronecker-factor trust regions (ACKTR). 
More recently, \citet{schulman2017proximal} introduced the Proximal Policy Optimization (PPO), a 
successor of TRPO which is easier to implement and has better sample complexity. For training policies for locomotion tasks with obstacles, \citet{heess2017emergence} proposed a distributed version of PPO.    
 
As a different direction towards sample efficiency, off-policy methods, such as Q-learning, were designed to use all the data collected from a system, regardless of the policies used for data generation. 
\citet{silver2014deterministic}, expanding on the work of \citet{degris2012off}, combined such ideas with the actor-critic framework into a method for training 
 deterministic policies, relying on exploratory policies. 
Later, \citet{lillicrap2015continuous} integrated this method with advances in deep Q-learning to obtain the Deep Deterministic Policy Gradient (DDPG) method. 

High variance of gradient estimation is not the only hurdle policy gradients methods need to surpass. 
Optimization problems occurring in RL are highly non-convex, leading many 
methods to find suboptimal local optima. To address this issue,  
\citet{haarnoja2017reinforcement} proposed the Soft Q-learning algorithm for 
learning multi-modal stochastic policies via entropy maximization, leading to better exploration in environments with multi-modal reward landscapes.  Recently, \citet{haarnoja2018soft} combined this idea with the actor-critic framework into the Soft Actor-Critic (SAC) algorithm, an off-policy actor-critic method in which the actor aims to maximize both the expected reward and the entropy of a stochastic policy. From a different direction,
 \citet{rajeswaran2017towards} 
used linear policies as a way of simplifying the search space. They used natural gradients, which are policy gradients adapted to the metric of the parameter space of the policy \cite{kakade2002natural}, to train linear policies 
for the MuJoCo locomotion tasks. 

While all these methods rely on exploration in the action space, there are model-free RL methods which perform 
exploration in the parameter space of the policies.  
Traditional finite difference gradient estimation for model-free RL uses coordinate aligned perturbations of policy weights 
and linear regression for measurement aggregation \cite{peters2008reinforcement}. Our method 
is based on finite differences along uniformly distributed directions; inspired by the derivative 
free optimization methods analyzed by \citet{nesterov2017random}, and similar to the Evolution Strategies 
algorithm \cite{salimans2017evolution}. The convergence of random search methods for derivative free optimization 
has been understood for several types of convex optimization \cite{agarwal2010optimal, bach2016highly, jamieson2012query, nesterov2017random}. \citet{jamieson2012query} offer an information theoretic lower bound for derivative free convex optimization and show that a coordinate based random search method achieves the lower bound with nearly optimal dependence on 
the dimension.

Although the efficiency of finite difference random search methods for derivative free convex optimization has been proven theoretically, 
these methods are perceived as inefficient when applied to nonconvex RL problems \cite{plappert2017parameter, peters2008reinforcement}. We offer evidence for the contrary. 


\section{ Problem setup}
\label{sec:prelims}
 
Solving problems in reinforcement learning requires finding policies for controlling dynamical systems with 
the goal of maximizing average reward on given tasks. Such problems can be abstractly formulated as  
\begin{align}
\label{eq:main_rl}
\max_{\theta \in \RR^d}\; \EE_\xi \left[ r(\pi_\theta, \xi) \right] \,,
\end{align}
where $\theta\in \RR^\statedim$ parametrizes a policy $\pi_\theta \colon \RR^\statedim \rightarrow \RR^\inputdim$. The random variable $\xi$ 
encodes the randomness of the environment, i.e., random initial states and stochastic transitions. 
The value $r(\pi_\theta, \xi)$ is the reward achieved by the policy $\pi_\theta$ on one 
trajectory generated from the system.
In general one could use stochastic policies $\pi_\theta$, but our proposed 
method uses deterministic policies. 

\subsection{Basic random search}\label{sec:brs}

Note that the problem formulation~\eqref{eq:main_rl} aims to optimize reward by directly optimizing over the policy parameters $\theta$. We consider methods which explore in the parameter space rather than the action space. This choice renders RL training equivalent to derivative-free optimization with noisy function evaluations. One of the simplest and oldest optimization methods for derivative-free optimization is \emph{random search}~\cite{matyas1965random}. Random search chooses a direction uniformly at random on the sphere in parameter space, and then optimizes the function along this direction.

A primitive form of random search simply computes a finite difference approximation along the random direction and then takes a step along this direction without using a line search. Our method ARS, described 
in Section~\ref{sec:alg}, is based precisely on this simple strategy.
For updating the parameters $\theta$ of a policy $\pi_\theta$, 
our method exploits update directions of the form:
\begin{align}
\label{eq:g_hat}
\frac{r(\pi_{\theta + \nu \delta}, \xi_1) - r(\pi_{\theta - \nu \delta}, \xi_2)}{\nu},
\end{align}
for two i.i.d. random variables $\xi_1$ and $\xi_2$, $\nu$ a positive real number, and $\delta$ a zero mean Gaussian vector.
It is known that such an update increment is an unbiased estimator of the gradient with respect to $\theta$ of $\EE_{\delta}\EE_\xi \left[r(\pi_{\theta + \nu \delta}, \xi)\right]$, a smoothed
version of the objective \eqref{eq:main_rl} which is close to the original objective \eqref{eq:main_rl} when $\nu$ is small \cite{nesterov2017random}. When the function evaluations are noisy, minibatches can be used to reduce the variance in this gradient estimate. The basic random search (BRS) algorithm is outlined in Algorithm~\ref{alg:ars_vone}. Evolution Strategies is version of this algorithm with several complicated algorithmic enhancements \cite{salimans2017evolution}. BRS is called Bandit Gradient Descent by \citet{flaxman2005online}. We note the many names for this algorithm, as it is at least 50 years old and has been rediscovered by a variety of different optimization communities.

\begin{center}
\begin{algorithm}[h!]
\begin{algorithmic}[1]
\STATE {\bf Hyperparameters:} step-size $\alpha$, number of directions sampled per iteration $N$, standard deviation of the exploration noise $\nu$
\STATE {\bf Initialize:} $\theta_0 = \mathbf{0}$, and $j = 0$.
\WHILE{ending condition not satisfied}
\STATE Sample $\delta_1, \delta_2, \ldots, \delta_N$ of the same size as $\theta_j$, with i.i.d. standard normal entries. 
\STATE Collect $2N$ rollouts of horizon $H$ and their corresponding rewards using the policies 
\begin{align*}
\pi_{j,k,+}(x) = \pi_{\theta_j + \nu \delta_k}(x) \quad \text{and} \quad \pi_{j,k,-}(x) = \pi_{\theta_j - \nu \delta_k}(x),
\end{align*} 
with $k \in \{1,2,\ldots, N\}$. 
\STATE Make the update step: \label{line:update_vone}
\begin{align*}
 \theta_{j + 1} = \theta_j + \tfrac{\alpha}{N} \sum_{k = 1}^{N}\left[r(\pi_{j,k, +}) - r(\pi_{j,k,-})\right]\delta_{k}\,.
\end{align*}
\STATE $j\leftarrow j + 1$.
\ENDWHILE
\end{algorithmic}
\caption{Basic Random Search (BRS)}. 
\label{alg:ars_vone}
 \end{algorithm}
\end{center} 

\subsection{An oracle model for RL}\label{sec:sample-complexity-setup}

We introduce and oracle model for RL to quantify the information about the system used by many RL methods. An RL algorithm can 
query the oracle by sending it a proposed policy $\pi_\theta$. Then, the oracle samples a random variable $\xi$, independent from the past, 
 and generates a trajectory from the system according to the policy $\pi_\theta$ and the randomness $\xi$. 
Then, the oracle returns to the RL algorithm a sequence of states, actions, and rewards $\{(s_t, a_t, r_t)\}_{t = 0}^{H - 1}$
which represent a trajectory generated from the system according to the policy $\pi_\theta$. One query is called an episode
or a rollout. The goal of RL algorithms is to approximately solve problem~\eqref{eq:main_rl} by making as few calls to the 
oracle as possible. The number of oracle queries needed for solving problem \eqref{eq:main_rl} is called
\emph{oracle complexity} or \emph{sample complexity}. Note that both policy gradient methods and finite difference methods 
can be implemented under this oracle model. Both approaches access the same information about the system: rollouts from fixed policies, and the associated states and rewards. The question then is whether one approach is making better use of this information than the other?

\section{Our proposed algorithm}
\label{sec:alg}

We now introduce three augmentations of BRS that build on successful heuristics employed in deep reinforcement learning. Throughout the rest of the paper we use $M$ to denote the parameters of policies because our method 
uses linear policies, and hence $M$ is a $\inputdim \times \statedim$ matrix. 

The first version of our method, ARS \vone{}, is obtained from BRS by scaling its update steps 
by the standard deviation of the rewards collected at each iteration (see  Line~\ref{line:update_vone} of Algorithm~\ref{alg:ars_vone}).
We motivate this scaling and offer intuition in Section~\ref{sec:sigmar}. As shown in Section~\ref{sec:mujoco}, 
\name{} \vone{} can train linear policies, which achieve the reward thresholds 
previously proposed in the literature, for the Swimmer-v1, Hopper-v1, 
HalfCheetah-v1, Walker2d-v1, and Ant-v1 tasks. 

However, \name{} \vone{} requires a larger number of episodes for training policies for these 
tasks, and it cannot train policies for the Humanoid-v1 task. To address these 
issues, in Algorithm~\ref{alg:lin-rl} we also propose \name{} \vtwo{}. ARS \vtwo{} 
trains policies which are linear maps of states normalized by a mean and standard deviation computed online.  
We explain further this procedure in Section~\ref{sec:non-isotropic}. 

To further enhance the performance of ARS \vone{} and ARS \vtwo{}, we introduce a third algorithmic enhancement, shown in Algorithm~\ref{alg:lin-rl} as ARS \vonet{} and ARS \vtwot{}. 
These versions of ARS can drop perturbation directions that yield the least improvement of the reward.
We motivate this algorithmic element in Section~\ref{sec:top_sampling}. 

\begin{center}
\begin{algorithm}[h!]
\begin{algorithmic}[1]
\STATE {\bf Hyperparameters:} step-size $\alpha$, number of directions sampled per iteration $N$, standard deviation of the exploration noise $\nu$,
number of top-performing directions to use $b$ ($b < N$ is allowed only for \vonet{} and \vtwot{})
\STATE {\bf Initialize:} $M_0 = \mathbf{0} \in \RR^{\inputdim \times \statedim}$, $\mu_0 = \mathbf{0} \in \RR^{\statedim}$, and $\Sigma_0 = \id_\statedim \in \RR^{\statedim \times \statedim}$, $j = 0$.
\WHILE{ending condition not satisfied}
\STATE Sample $\delta_1, \delta_2, \ldots, \delta_N$ in $\R^{\inputdim \times \statedim}$ with i.i.d. standard normal entries. 
\STATE Collect $2N$ rollouts of horizon $H$ and their corresponding rewards using the $2N$ policies \label{line:collection}
\begin{align*}
  &\textbf{V1: }\begin{cases} \pi_{j,k,+}(x) = (M_j + \nu \delta_k)x \\
                \pi_{j,k,-}(x) = (M_j - \nu \delta_k)x 
    \end{cases} \\
&\textbf{V2: } \begin{cases} \pi_{j,k,+}(x) = (M_j + \nu \delta_k) \diag\left(\Sigma_j\right)^{-\sfrac{1}{2}}(x - \mu_j)\\
 \pi_{j,k,-}(x) = (M_j - \nu \delta_k) \diag(\Sigma_j)^{-\sfrac{1}{2}}(x - \mu_j)
 \end{cases}\\
\end{align*}
for $k\in \{1,2,\ldots, N\}$. 
\STATE Sort the directions $\delta_k$ by $\max\{r(\pi_{j,k,+}), r(\pi_{j,k,-})\}$, denote by $\delta_{(k)}$ the $k$-th largest direction,
and by $\pi_{j,(k),+}$ and $\pi_{j,(k),-}$ the corresponding policies.  
\STATE Make the update step: \label{line:update_step}
\begin{align*}
M_{j + 1} = M_j + \tfrac{\alpha}{b \sigma_R} \sum_{k = 1}^{b}\left[r(\pi_{j,(k), +}) - r(\pi_{j,(k),-})\right]\delta_{(k)}, 
\end{align*}
where  $\sigma_R$ is the standard deviation of the $2b$ rewards used in the update step. 
\STATE \vtwo{} : Set $\mu_{j + 1}$, $\Sigma_{j + 1}$ to be the mean and covariance of the $2 N H(j + 1)$ states encountered from the start of training.\footnotemark \label{line:update_stats}
\STATE $j\leftarrow j + 1$
\ENDWHILE
\end{algorithmic}
\caption{Augmented Random Search (\name{}): four versions \vone{}, \vonet{}, \vtwo{} and \vtwot{}}. 
\label{alg:lin-rl}
 \end{algorithm}
\end{center}

\subsection{Scaling by the standard deviation $\sigma_R$}
\label{sec:sigmar}

As the training of policies progresses, random search in the parameter space of policies can lead 
to large variations in the rewards observed across iterations. 
As a result, it is difficult to choose a fixed step-size $\alpha$ which does not 
allow harmful changes between large and small steps. 
\citet{salimans2017evolution} address this issue by transforming the rewards into 
rankings and then using the adaptive optimization algorithm Adam for computing the update step. 
Both of these techniques change the direction of the updates, obfuscating the behavior of the algorithm 
and making it difficult to ascertain the objective Evolution Strategies is actually optimizing. 

\footnotetext{Of course, we implement this in an efficient way that does not require the storage of all the states. Also, we only keep track of the diagonal of $\Sigma_{j + 1}$. Finally, to ensure that the ratio $0/0$ is treated as $0$, if a 
diagonal entry of $\Sigma_j$ is smaller than $10^{-8}$ we make it equal to $+\infty$.}
 
\begin{figure}
\centerline{\includegraphics[width=0.4\textwidth]{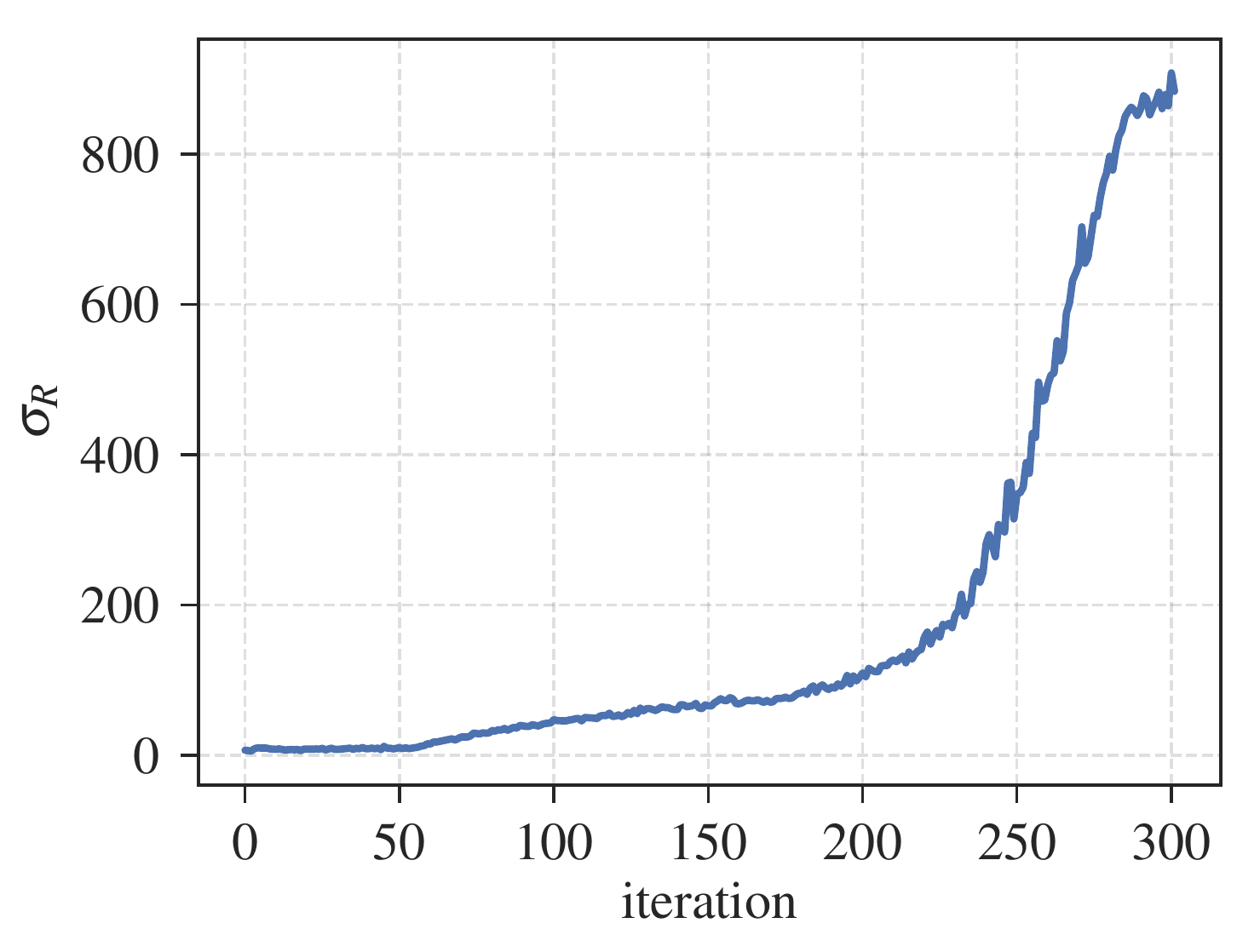}}
\captionsetup{width=.7\linewidth}
\caption{Showing the standard deviation $\sigma_R$ of the rewards collected at each iteration, while training Humanoid-v1.}
\label{fig:sigmar}
\end{figure}

To address the large variations of the differences $r(\pi_{M + \nu \delta}) - r(\pi_{M - \nu \delta})$,
we scale the update steps by 
the standard deviation $\sigma_R$ of the $2N$ rewards collected at each iteration (see Line~\ref{line:update_step} of Algorithm~\ref{alg:lin-rl}).
To understand the effect of scaling by $\sigma_R$, we plot 
standard deviations $\sigma_R$ obtained during training a policy for the Humanoid-v1 model in Figure~\ref{fig:sigmar}. 
 The standard deviations $\sigma_R$ have an increasing trend as training progresses. This behavior 
occurs because perturbations of the policy weights at high rewards can cause Humanoid-v1 to fall 
early, yielding large variations in the rewards collected. Therefore, without 
scaling by $\sigma_R$, our method at iteration $300$ would be taking steps which are a thousand times 
larger than in the beginning of training. The same effect of scaling by $\sigma_R$ could probably be obtained
by tuning a step-size schedule. However, our goal was to minimize the amount of tuning required, and 
thus we opted for the scaling by the standard deviation. 
 
\subsection{Normalization of the states}
\label{sec:non-isotropic}

The normalization of states used by \vtwo{} is akin to data whitening used in regression tasks, and intuitively it ensures that policies put equal weight on the different components of the states. To gain intuition for why this might help, suppose that a state coordinate only takes values in the range $[90,100]$ while another state component takes values in the range $[-1,1]$.  Then, small changes in the control gain with respect to the first state coordinate would lead to larger changes in the actions then the same sized changes with respect to the second state component. Hence, whitening allows the isotropic exploration of random search to have equal influence over the various state components.

Previous work has also implemented such state normalization for fitting a neural network model for several MuJoCo environments~\cite{nagabandi2017neural}. A similar normalization is used by ES as part of the virtual batch normalization of the neural network policies \cite{salimans2017evolution}.

In the case of ARS, the state normalization can be seen as a form of non-isotropic exploration
in the parameter space of linear policies. In particular, for policy weights $M$ and a perturbation direction $\delta$ 
we have 
\begin{align}
\label{eq:nonisotropic}
\left(M + \nu\delta \right)\diag(\Sigma)^{-\sfrac{1}{2}}(x - \mu) = \left(\tilde{M} + \nu\delta \diag(\Sigma)^{-\sfrac{1}{2}} \right)(x - \mu), 
\end{align}
where $\tilde{M} = M \diag(\Sigma)^{-\sfrac{1}{2}}$. 
 
The main empirical motivation for version 2 of our method comes from the Humanoid-v1 task.
We were not able to train a linear policy for this task without the normalization of the 
states described in Algorithm~\ref{alg:lin-rl}. 
Moreover, the measured sample complexity of \name{} \vtwo{} is better on the other MuJoCo locomotion tasks as well, as shown in Section~\ref{sec:mujoco}. On the other hand, we note that ARS \vtwo{} is impractical for the 
Linear Quadratic Regulator problem, discussed in Section~\ref{sec:lqr}, because the size of the states 
grows exponentially fast as a function of the trajectory length when the policy does not stabilize the system.  

\subsection{Using top performing directions}
\label{sec:top_sampling}

In Section~\ref{sec:mujoco} we show that ARS \vtwo{}  matches or exceeds state-of-the-art performance 
on the tasks Swimmer-v1, Hopper-v1, HalfCheetah-v1 and Humanoid-v1. However, for training the Walker2d-v1 and Ant-v1 models, 
\name{} \vtwo{} requires two to three times more rollouts than competing methods. 

To improve the performance of ARS \vone{} and \vtwo{} we propose \name{} \vonet{} and \vtwot{}. 
In the update steps used by \name{} \vone{} and \vtwo{} each perturbation direction $\delta$ is 
weighted by the difference of the rewards $r(\pi_{j,k, +})$ and $r(\pi_{j,k, -})$. These two rewards are 
the obtained from two queries to the oracle described in Section~\ref{sec:prelims}, using the policies 

\begin{align*}
\pi_{j,k,+}(x) = (M_j + \nu \delta_k) \diag\left(\Sigma_j\right)^{-\sfrac{1}{2}}(x - \mu_j) \quad \text{and} \quad \pi_{j,k,-}(x) = (M_j - \nu \delta_k) \diag(\Sigma_j)^{-\sfrac{1}{2}}(x - \mu_j).
\end{align*}
If $r(\pi_{j,k, + }) > r(\pi_{j,k,-})$, the update steps of ARS \vone{} and \vtwo{} push the policy weights $M_j$ in
the direction of $\delta_k$. 
If $r(\pi_{j,k, + }) < r(\pi_{j,k,-})$, the update steps of ARS \vone{} and \vtwo{} push the policy weights $M_j$ in
the direction of $-\delta_k$. 
However, since $r(\pi_{j,k, +})$ and $r(\pi_{j,k,-})$ are noisy evaluations of the performance of the policies parametrized by  $M_j + \nu \delta_k$ and $M_j - \nu \delta_k$, ARS \vone{} and \vtwo{} might push the weights $M_j$ in the direction $\delta_k$ even when $-\delta_k$ is better,
or vice versa. 
Moreover, there can be perturbation directions $\delta_k$ such that updating the policy weights $M_j$ 
in either the direction $\delta_k$ or $-\delta_k$ would lead to sub-optimal performance. 

For example, the rewards $r(\pi_{j,k, +})$ and $r(\pi_{j,k,-})$ being both small compared to other 
observed rewards might suggest that moving $M_j$ in either the direction $\delta_k$ or $-\delta_k$
would decrease the average reward. To address these issues, in ARS \vonet{} and \vtwot{} we propose to  
order decreasingly the perturbation directions $\delta_k$, according to $\max\{r(\pi_{j,k,+}), r(\pi_{j,k,-})\}$, and then use only the top $b$ directions for updating the policy weights (see Line~\ref{line:update_step} of Algorithm~\ref{alg:lin-rl}).

This algorithmic enhancement intuitively improves the update steps of ARS because with it the update steps are an average
 over directions that obtained high rewards. However, without theoretical investigation we cannot be certain of the effect of using 
this algorithmic enhancement (i.e. choosing $b < N$). When $b = N$ versions  \vonet{} and \vtwot{} are equivalent to versions \vone{} and \vtwo{}. 
Therefore, it is certain that after tuning the hyperparameters of ARS \vonet{} and \vtwot{}, they will not perform any worse than ARS 
\vone{} and \vtwo{}. 
In Section~\ref{sec:mujoco} we show that ARS \vtwot{} exceeds or matches state-of-the-art performance
on all the MuJoCo locomotion tasks included in the OpenAI gym.

\subsection{Comparison to \citet{salimans2017evolution}}
\label{sec:es}

\name{} simplifies the Evolution Strategies of \citet{salimans2017evolution} in several ways:
\begin{itemize}
\item ES feeds the gradient estimate into the Adam algorithm.
\item Instead of using the actual reward values $r(\theta \pm \sigma \epsilon_i)$, ES transforms the rewards into rankings and uses the ranks to compute update steps. The rankings are used to make training more robust. Instead, our method scales the update steps by the standard deviation of the rewards. 
\item ES bins the action space of the Swimmer-v1 and Hopper-v1 to encourage exploration. Our method surpasses ES without such binning. 
\item ES relies on policies parametrized by neural networks with virtual batch normalization, while we show that \name{} 
achieves state-of-the-art performance with linear policies. 
\end{itemize}


\section{Experimental results}
\label{sec:empirical}

\subsection{Implementation details}
\label{sec:implementation_details}

We implemented a parallel version of Algorithm~\ref{alg:lin-rl} using the Python library Ray~\cite{moritz2017ray}.
To avoid the computational bottleneck of communicating perturbations $\delta$, we created a shared noise table 
which stores independent standard normal entries. Then, instead of communicating perturbations $\delta$,
the workers communicate indices in the shared noise table. This approach has been used in the implementation 
of ES by \citet{moritz2017ray} and is similar to the approach proposed by~\citet{salimans2017evolution}. 
Our code sets the random seeds for the random generators of all the workers and for all copies of the OpenAI Gym environments held by the workers. 
All these random seeds are distinct and are a function of a single integer to which we refer as \emph{the random seed}. 
Furthermore, we made sure that the states and rewards produced during the evaluation rollouts were not used in any form 
during training.   

\subsection{Results on the MuJoCo locomotion tasks}
\label{sec:mujoco}

We evaluate the performance of \name{} on the MuJoCo locomotion tasks included in the OpenAI Gym-v$0.9.3$~\cite{1606.01540, todorov2012mujoco}. 
The OpenAI Gym provides benchmark reward functions for the different MuJoCo locomotion tasks. 
We used these default reward functions for evaluating the performance of the linear policies trained with \name{}. 
The reported rewards obtained by a policy were averaged over $100$ independent rollouts. 

For the Hopper-v1, Walker2d-v1, Ant-v1, and Humanoid-v1 tasks the default reward functions include a survival bonus, which rewards RL agents with a constant
reward at each timestep, as long as a termination condition (i.e., falling over) has not been reached. 
For example, the environment Humanoid-v1 awards a reward of $5$ at each time steps, as long as the Humanoid model does not fall.
Hence, if the Humanoid model stands still for $1000$ timesteps, it will receive a reward of $5000$ minus a small penalty for 
the actions used to maintain a vertical position. Furthermore, if the Humanoid falls forward at then end of a rollout, it will 
receive a reward higher than $5000$.

It is common practice to report the sample complexity of an RL method by showing the number of episodes required to reach a reward threshold \cite{gu2016q, rajeswaran2017towards, salimans2017evolution}.  
For example, \citet{gu2016q} chose a threshold of $2500$, while \citet{rajeswaran2017towards} chose a threshold of $5280$.
However, given the survival bonus awarded to Humanoid-v1, we do not believe these reward thresholds are meaningful for locomotion. 
In Table~\ref{table:sample_complexity} and Section~\ref{sec:computation} we use a reward threshold of $6000$ to evaluate the performance of \name{} on the Humanoid-v1 task, the threshold also used by \citet{salimans2017evolution}. 

The survival bonuses awarded by the OpenAI gym discourage the exploration of 
policies that cause falling early on, which is needed for the discovery of policies that achieve locomotion.
These bonuses cause ARS to find policies which make the MuJoCo models 
stand still for a thousand timesteps; policies which are likely local optima. 
These bonuses were probably included in the reward functions to help the training of stochastic policies since 
such policies cause constant movement through stochastic actions. 
To resolve the local optima problem for training deterministic policies, we subtracted the survival bonus 
from the rewards outputted by the OpenAI gym during training.
For the evaluation of trained policies we used the default reward functions.  

We first evaluated the performance of \name{} on three random seeds after hyperparameter tuning. 
Evaluation on three random seeds is widely adopted in the literature and hence we wanted 
to put ARS on equal footing with competing methods. 
Then, we evaluated the performance of \name{} on $100$ random seeds for a thorough estimation of performance. 
Finally, we also evaluated the sensitivity of our method to changes of the hyperparameters.

\paragraph{Three random seeds evaluation:} We compared the different versions of \name{} against the following methods: Trust Region Policy Optimization (TRPO), Deep Deterministic Policy Gradient (DDPG), Natural Gradients (NG), Evolution Strategies (ES), Proximal Policy Optimization (PPO), Soft Actor Critic (SAC), Soft Q-Learning (SQL), A2C, and the Cross Entropy Method (CEM). For the performance of these methods we used values reported by \citet{rajeswaran2017towards}, \citet{salimans2017evolution}, \citet{schulman2017proximal}, and \citet{haarnoja2018soft}.

\citet{rajeswaran2017towards} and \citet{schulman2017proximal} evaluated the performance of RL algorithms on three random seeds, while \citet{salimans2017evolution} and \citet{haarnoja2018soft} used six and five random seeds respectively.
To all methods on equal footing, for the evaluation of \name{}, we sampled three random seeds uniformly from the interval $[0, 1000)$ and fixed them. For each of the six popular MuJoCo locomotion tasks we chose a grid of hyperparameters\footnote{Recall that \name{} \vone{} and \vtwo{} take in only three hyperparameters:  the step-size $\alpha$, the number of perturbation directions $N$, and scale of the perturbations $\nu$. \name{} \vonet{} and \vtwot{} take in an additional hyperparameter, the number of 
top directions used $b$ ($b \leq N$).}, shown in Appendix~\ref{sec:parameters}, and for each set of hyperarameters we ran \name{} \vone{}, \vtwo{}, \vonet{}, and \vtwot{} three times, once for each of the three fixed random seeds.

\begin{figure}[h]
\centering
\begin{subfigure}[b]{0.33\textwidth}
\centerline{\includegraphics[width=1.\columnwidth]{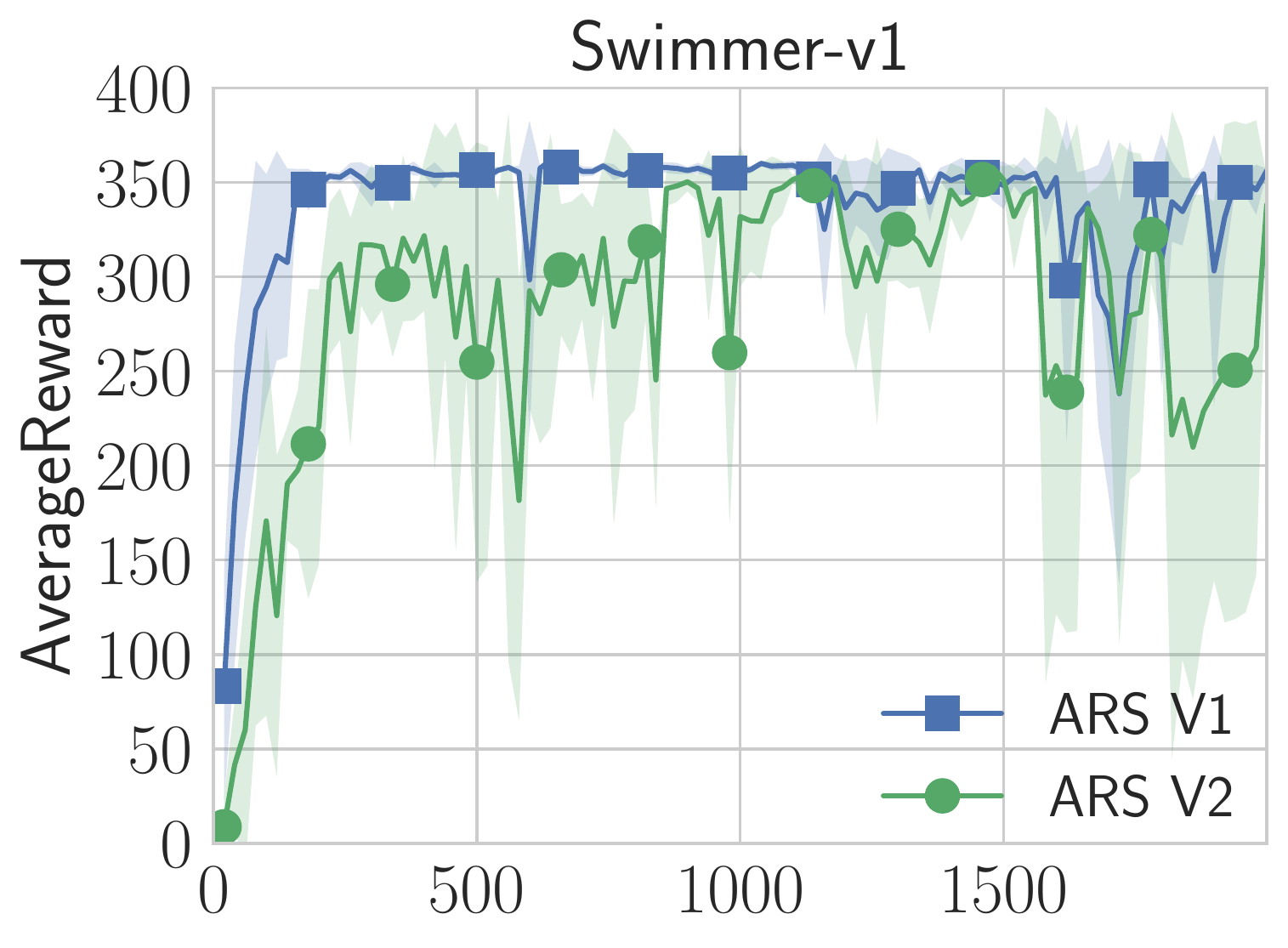}}
\end{subfigure}
\begin{subfigure}[b]{0.33\textwidth}
\centerline{\includegraphics[width=1.\columnwidth]{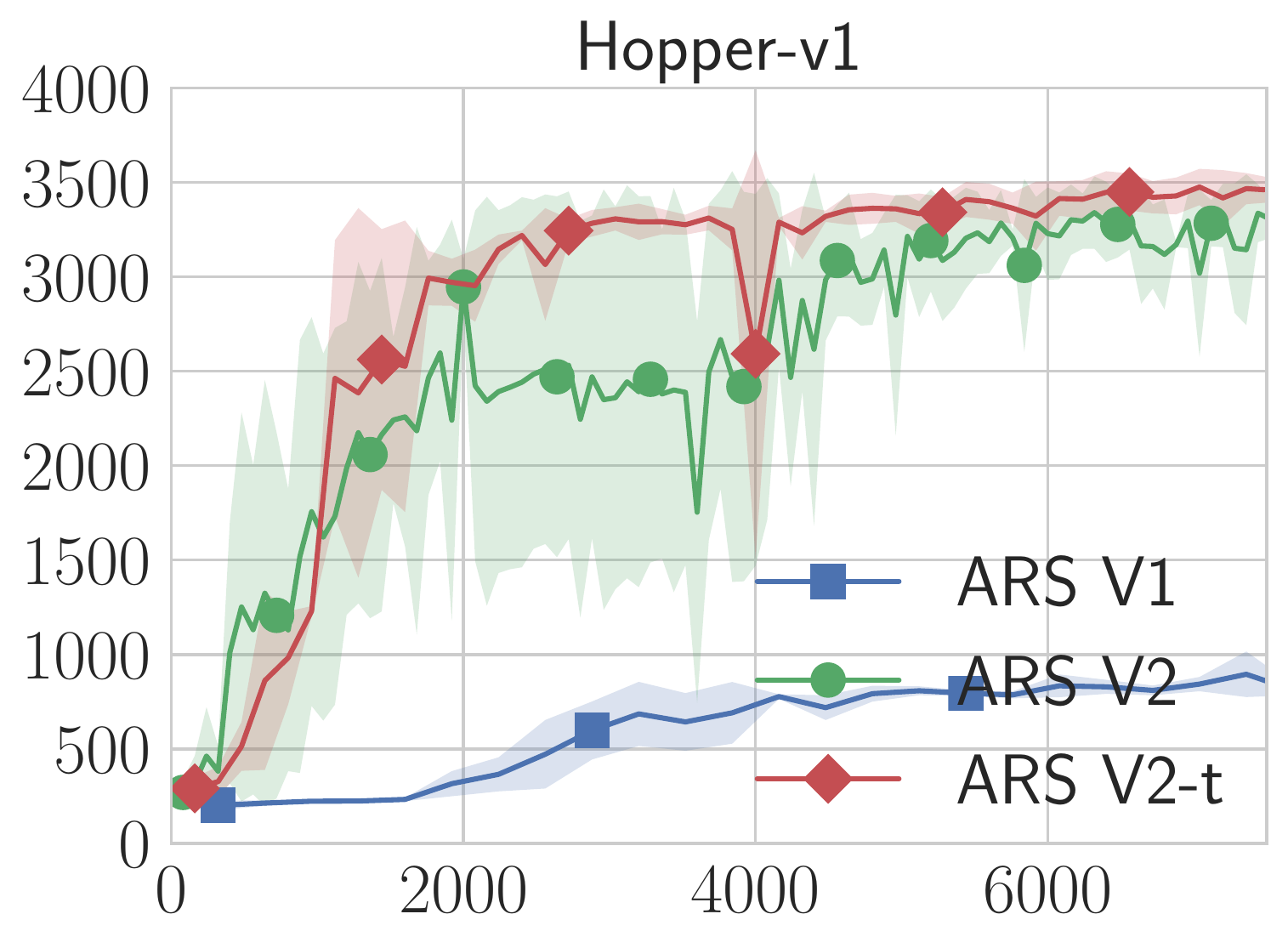}}
\end{subfigure}
\begin{subfigure}[b]{0.33\textwidth}
\centerline{\includegraphics[width=1.\columnwidth]{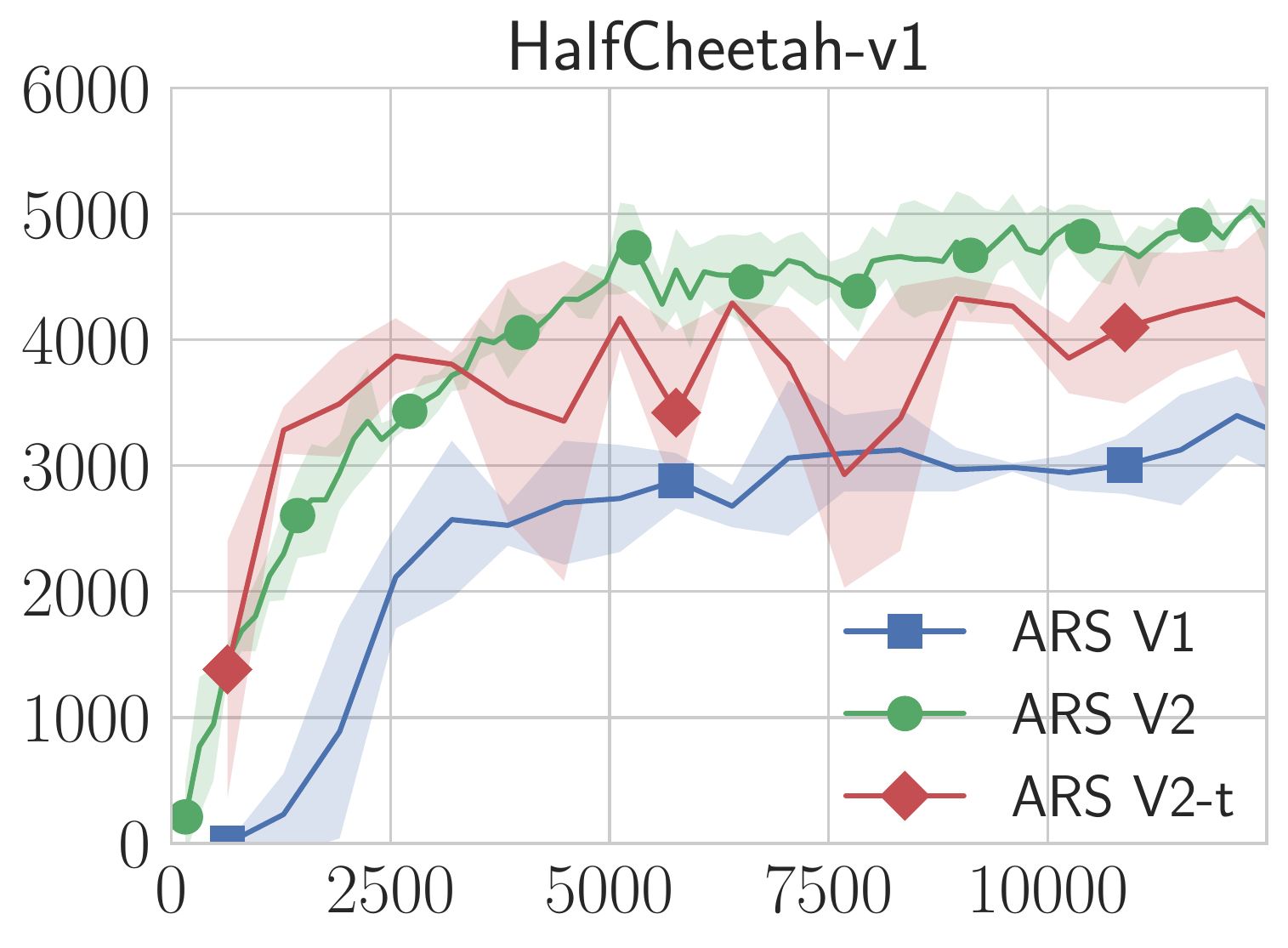}}
\end{subfigure}
\begin{subfigure}[b]{0.33\textwidth}
\centerline{\includegraphics[width=1.\columnwidth]{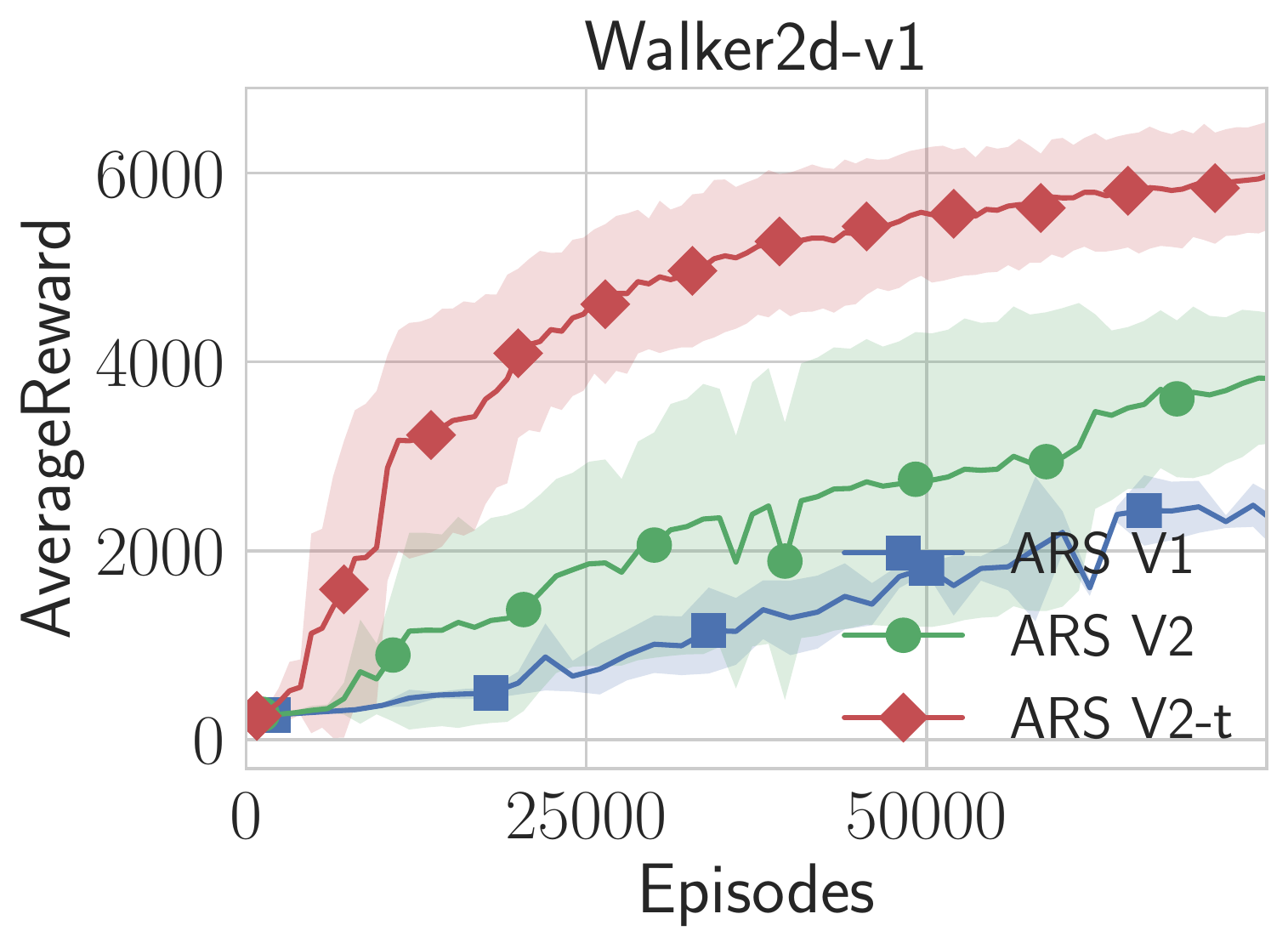}}
\end{subfigure}
\begin{subfigure}[b]{0.33\textwidth}
\centerline{\includegraphics[width=1.\columnwidth]{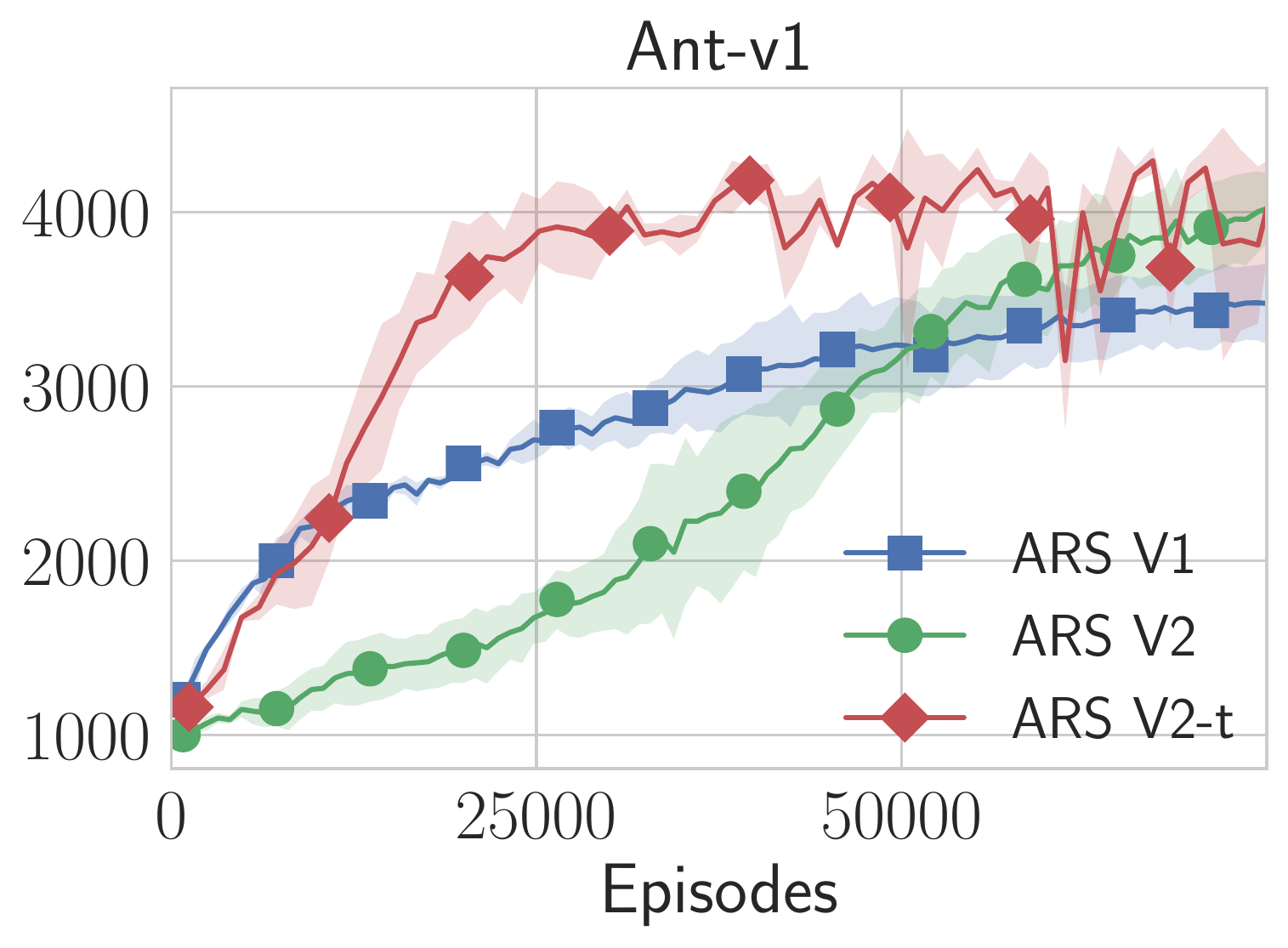}}
\end{subfigure}
\begin{subfigure}[b]{0.33\textwidth}
\centerline{\includegraphics[width=1.\columnwidth]{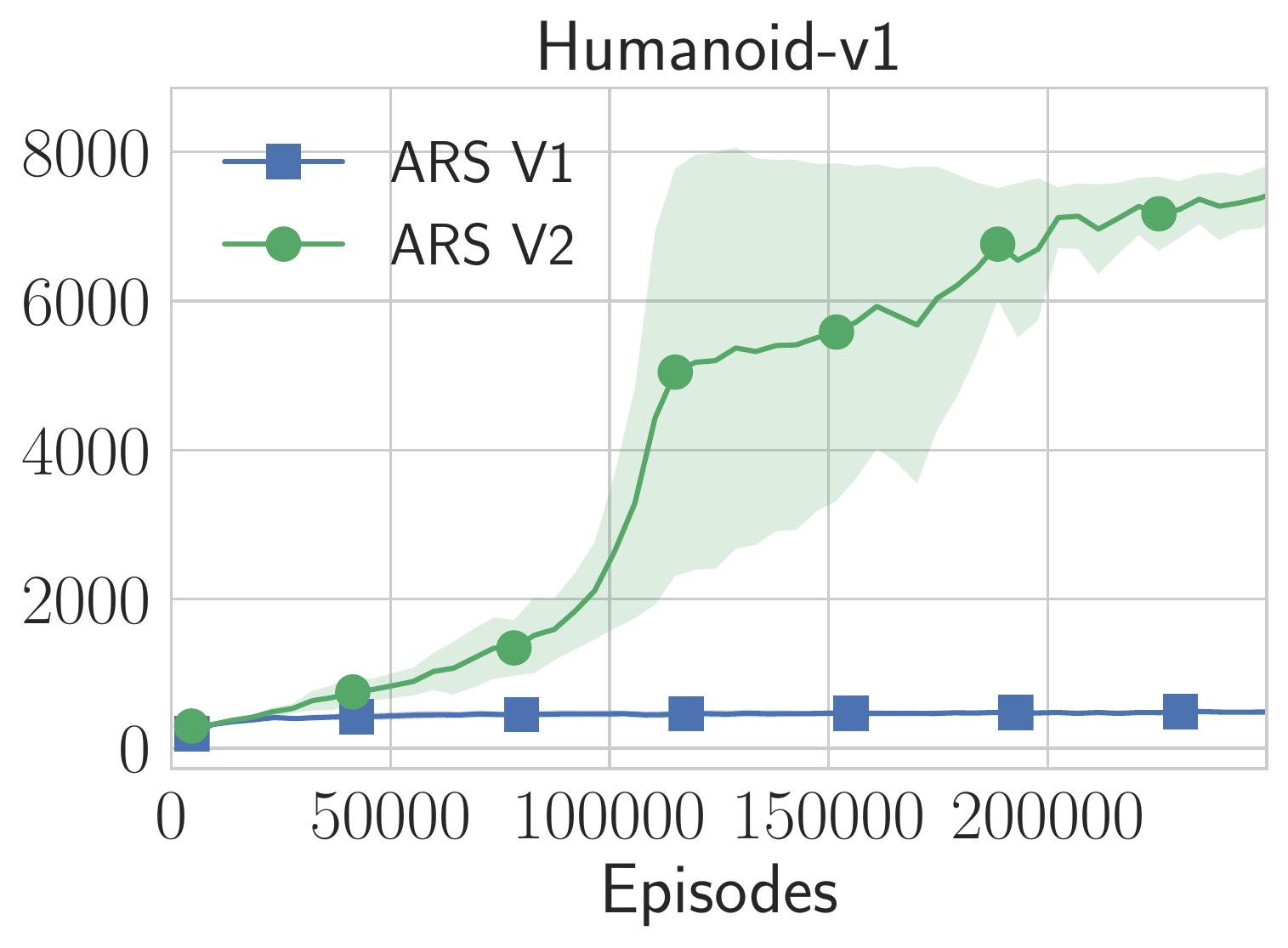}}
\end{subfigure}
\caption{An evaluation of four versions of \name{} on the MuJoCo locomotion tasks. The training 
curves are averaged over three random seeds, and the shaded region shows the standard deviation.
ARS \vtwot{} is only shown for the tasks to which it offered an improvement over ARS \vtwo{}.}
\label{fig:3_seeds}
\end{figure}

Table~\ref{table:sample_complexity} shows the average number of episodes required by \name{}, NG, and TRPO
to reach a prescribed reward threshold, using the values reported by \citet{rajeswaran2017towards} for NG and TRPO. 
For each version of \name{} and each MuJoCo task we chose the hyperparameters which minimize the average number 
of episodes required to reach the reward threshold. The corresponding training curves of \name{} are shown in Figure~\ref{fig:3_seeds}. 
For all MuJoCo tasks, except Humanoid-v1, we used the
same reward thresholds as \citet{rajeswaran2017towards}. Our choice to increase the reward threshold for Humanoid-v1 
is motivated by the presence of the survival bonuses, as discussed in Section~\ref{sec:implementation_details}. 

\begin{table}[h]
\centering 
\resizebox{\textwidth}{!}{\begin{tabular}{cc|cccc|ccc}
&& \multicolumn{7}{c}{{\bf Average \# episodes to reach reward threshold}} \\
\hline
\textbf{Task} & \textbf{Threshold} & & \multicolumn{2}{c}{{\bf ARS}} & & \textbf{NG-lin} & \textbf{NG-rbf} & \textbf{TRPO-nn} \\
\hline 
&& \textbf{\vone{}} & \textbf{\vone{}-t} & \textbf{\vtwo{}} & \textbf{\vtwo{}-t} &&& \\
Swimmer-v1 & $325 $  & $100$   & $100$ & $427$  & $427$ &  $1450$ & $1550$ & N/A \footnotemark \\
Hopper-v1  &  $3120$  & $89493$ & $51840$ & $3013$ & $1973$  & $13920$ & $8640$ &$10000$ \\
HalfCheetah-v1 &  $3430$  & $10240$ & $8106$ & $2720$ & $1707$ & $11250$ & $6000$ &$4250$\\
Walker2d-v1  &  $4390$  & $392000$ & $166133$ & $89600$ & $24000$ & $36840$ & $25680$ & $14250$\\
Ant-v1     &  $3580$  & $101066$ & $58133$ & $60533$ & $20800$ & $39240$ & $30000$ & $73500$\\
Humanoid-v1 & $6000$ & N/A  &  N/A  & $142600$ & $142600$ & $\approx\!\! 130000$ &  $\approx\!\! 130000$ & UNK\footnotemark \\
\hline
\end{tabular}}
\caption{A comparison of ARS, NG, and TRPO on the MuJoCo locomotion tasks.    
For each task we show the average number of episodes required to achieve a prescribed reward threshold, averaged over three random seeds. 
We estimated the number of episodes required by NG to reach a reward of $6000$ for Humanoid-v1 based on the learning curves presented by \citet{rajeswaran2017towards}.}
\label{table:sample_complexity}




 


\end{table}

\addtocounter{footnote}{-2} 
\stepcounter{footnote}\footnotetext{N/A means that the method did not reach the reward threshold.}
\stepcounter{footnote}\footnotetext{UNK stands for unknown.}

Table~\ref{table:sample_complexity} shows that \name{} \vone{} can train policies for all the MuJoCo locomotion tasks except Humanoid-v1, which  
 is successfully solved by \name{} \vtwo{}. 
Secondly, we note that \name{} \vtwo{} reaches the prescribed thresholds for the Swimmer-v1, Hopper-v1, and HalfCheetah-v1 tasks 
faster than NG or TRPO, and matches the performance of NG on the Humanoid-v1 task. On the Walker2d-v1 and Ant-v1 tasks \name{} \vtwo{} 
is outperformed by NG. Nonetheless, we note that \name{} \vtwot{} surpasses the performance of NG on these two tasks. 
Although TRPO hits the reward threshold for Walker2d-v1 faster than \name{}, we will see that in other metrics \name{} surpasses TRPO. 

Table~\ref{table:ppo} shows the maximum reward achieved by \name{}\footnote{We explain our methodology for computing this value for \name{} in Appendix~\ref{sec:timestep_comparison}.}, PPO, A2C, CEM, and TRPO after one million timesteps of the simulator have been collected, averaged over the three fixed 
random seeds. The hyperparameters were chosen based on the same evaluations performed for Table~\ref{table:sample_complexity} and Figure~\ref{fig:3_seeds}. 
\citet{schulman2017proximal} did not report performance of PPO, A2C, CEM, and TRPO on the Ant-v1 and Humanoid-v1 tasks of the OpenAI gym. 
Table~\ref{table:ppo} shows that \name{} surpasses these four methods on the Swimmer-v1, Hopper-v1, and HalfCheetah-v1 tasks. 
On the Walker2d-v1 task PPO achieves a higher average maximum reward than \name{}, while \name{} achieves a similar maximum reward to A2C, CEM, and TRPO. 

\begin{table}[h]
\centering 
\begin{tabular}{cc|ccccc}
& & \multicolumn{5}{c}{{\bf Maximum average reward after \# timesteps}}\\
 \hline
\textbf{Task} & \textbf{\# timesteps} & \textbf{\name{}} & \textbf{PPO} & \textbf{A2C} & \textbf{CEM} & \textbf{TRPO} \\
\hline 
Swimmer-v1     & $10^6$ & $361$ & $\approx \! 110$ & $\approx \! 30$ & $\approx \! 0$ & $\approx \! 120$\\
Hopper-v1  & $10^6$ &  $3047$ & $\approx \! 2300$ & $\approx\! 900$ &$\approx \! 500$ &  $\approx \! 2000$\\
HalfCheetah-v1 & $10^6$ & $2345$ & $\approx\! 1900$ & $\approx \! 1000$ & $\approx \! - 400$ &  $\approx \! 0$ \\
Walker2d-v1  & $10^6$ &  $894$ & $\approx \! 3500$ & $\approx \! 900$ & $\approx \! 800$  & $\approx \! 1000$\\
\hline
\end{tabular}
\caption{
A comparison of ARS, PPO, A2C, CEM, and TRPO on the MuJoCo locomotion tasks.    
For each task we show the maximum rewards achieved after a prescribed number of simulator timesteps have been used, averaged over three random seeds. The values for PPO, A2C, CEM, and TRPO were approximated based on the figures presented by \citet{schulman2017proximal}.}

\label{table:ppo}

\end{table}

Table~\ref{table:sac} shows the maximum reward achieved by \name{}, SAC, DDPG, SQL, and TRPO after a prescribed number of simulator timesteps have been collected. The hyperparameters for ARS were chosen based on the same evaluations performed for Table~\ref{table:sample_complexity} and Figure~\ref{fig:3_seeds}. 
Table~\ref{table:sac} shows that \name{} surpasses SAC, DDPG, SQL, and TRPO on the Hopper-v1 and Walker2d-v1 tasks, and that \name{} 
is surpassed by SAC, DDPG, and SQL on the HalfCheetah-v1 taks. However, ARS performs better than TRPO on this task. 
On the Ant-v1 task, \name{} is surpassed by SAC and performs similarly to SQL, but it outperforms DDPG and TRPO. 
We did not include values for Swimmer-v1 and Humanoid-v1 because \citet{haarnoja2018soft} did not use the OpenAI versions of these tasks 
for evaluation. Instead, they evaluated SAC on the rllab version \cite{duan2016benchmarking} of these tasks. 
The authors indicated that Humanoid-v1 is more challenging for SAC than the rllab version because of the parametrization of the states used by the OpenAI gym, 
and that Swimmer-v1 is more challenging because of the reward function used. 

\begin{table}[h]
\centering 
\begin{tabular}{cc|ccccc}
& & \multicolumn{5}{c}{{\bf Maximum average reward after \# timesteps}}\\
\hline
\textbf{Task} & \textbf{\# timesteps} & \textbf{\name{}} & \textbf{SAC} & \textbf{DDPG} & \textbf{SQL} & \textbf{TRPO} \\
\hline 
Hopper-v1  & $2.00 \cdot 10^6$ &  $3306$ & $\approx \! 3000$ & $\approx\! 1100$ &$\approx \! 1500$ &  $\approx \! 1250$\\
HalfCheetah-v1 & $1.00 \cdot 10^7$ & $5024$ & $\approx\! 11500$ & $\approx \! 6500$ & $\approx \! 8000$ &  $\approx \! 1800$ \\
Walker2d-v1  & $5.00 \cdot 10^6$ &  $4205$ & $\approx \! 3500$ & $\approx \! 1600$ & $\approx \! 2100$  & $\approx \! 800$\\
Ant-v1     & $1.00 \cdot 10^7$ & $2072$ & $\approx \! 2500$ & $\approx \! 200$ & $\approx \! 2000$ & $\approx \! 0$\\
\hline
\end{tabular}
\caption{
A comparison of ARS, SAC, DDPG, SQL, and TRPO on the MuJoCo locomotion tasks.    
For each task we show the maximum rewards achieved after a prescribed number of simulator timesteps have been used.
The values for ARS were averaged over three random seeds.
 The values for SAC, DDPG, SQL, and TRPO were approximated based on the figures presented by \citet{haarnoja2018soft}, who evaluated these methods on five random seeds.}
\label{table:sac}

\end{table}

Table~\ref{table:es} shows the number of timesteps required by \name{} to reach a prescribed reward threshold, 
averaged over the three fixed random seeds. 
The hyperparameters were chosen based on the same evaluations performed for Table~\ref{table:sample_complexity} and Figure~\ref{fig:3_seeds}. 
We compare ARS to ES and TRPO. For these two methods we show the values reported by \citet{salimans2017evolution}, who used six random seeds for evaluation. 
\citet{salimans2017evolution} do not report sample complexity results for the Ant-v1 and Humanoid-v1 tasks. 
Table~\ref{table:es} shows that TRPO requires fewer timesteps
than \name{} to reach the prescribed reward threshold on Walker2d-v1. However, we see that \name{} requires fewer
timesteps than ES and TRPO on the Swimmer-v1, Hopper-v1, and HalfCheetah-v1 tasks.

\begin{table}[h]
\centering 
\begin{tabular}{cc|ccc}
& & \multicolumn{3}{c}{{\bf Average \# timesteps to hit Th.}}\\
\hline
\textbf{Task} & \textbf{Threshold} & \textbf{\name{}} & \textbf{ES} & \textbf{TRPO}\\
\hline 
Swimmer-v1 & $128.25$ & $6.00 \cdot 10^4$ & $1.39 \cdot 10^6$ & $4.59 \cdot 10^6$ \\
Hopper-v1  & $3403.46$ &  $2.00 \cdot 10^6$ & $3.16 \cdot 10^7$ & $4.56 \cdot 10^6$ \\
HalfCheetah-v1 & $2385.79$ & $5.86 \cdot 10^5$ & $2.88 \cdot 10^6$ & $5.00 \cdot 10^6$ \\
Walker2d-v1  & $3830.03$ &  $8.14 \cdot 10^6$ & $3.79 \cdot 10^7$ & $4.81 \cdot 10^6$ \\
\hline
\end{tabular}
\caption{A comparison of \name{} and ES and TRPO methods on the MuJoCo locomotion tasks. For each task we show the average number of timesteps required by \name{} to reach a prescribed reward threshold, averaged over three random seeds. For Swimmer-v1 we used \name{} \vone{}, while for the other tasks we used \name{} \vtwot{}. The values for ES and TRPO have been averaged over six random seeds and are taken from \cite{salimans2017evolution}. \citet{salimans2017evolution} did not evaluate on Ant-v1 and they did not specify the exact number of timesteps required to train Humanoid-v1.}
\label{table:es}

\end{table}

\paragraph{A hundred seeds evaluation:} Evaluating ARS on three random seeds shows that
overall our method is more sample efficient than the NG, ES, DDPG, PPO, SAC, SQL, A2C, CEM, and TRPO methods on the
MuJoCo locomotion tasks. 
However, it is well known that RL algorithms exhibit high training variance~\cite{islam2017reproducibility, henderson2017deep}. 

For a thorough evaluation, we sampled $100$ distinct random seeds uniformly at random from the interval 
$[0,10000)$. Then, using the hyperparameters selected for Table~\ref{table:sample_complexity} and Figure~\ref{fig:3_seeds},
we ran \name{} for each of the six MuJoCo locomotion tasks and the $100$ random seeds. 
Such a thorough evaluation was feasible only because \name{} has a small computational footprint, as discussed
in Section~\ref{sec:computation}.
 
\begin{figure}[h]
\caption*{\bf Average reward evaluated over 100 random seeds, shown by percentile}
\centering
\begin{subfigure}[b]{0.32\textwidth}
\centerline{\includegraphics[height=3.5cm]{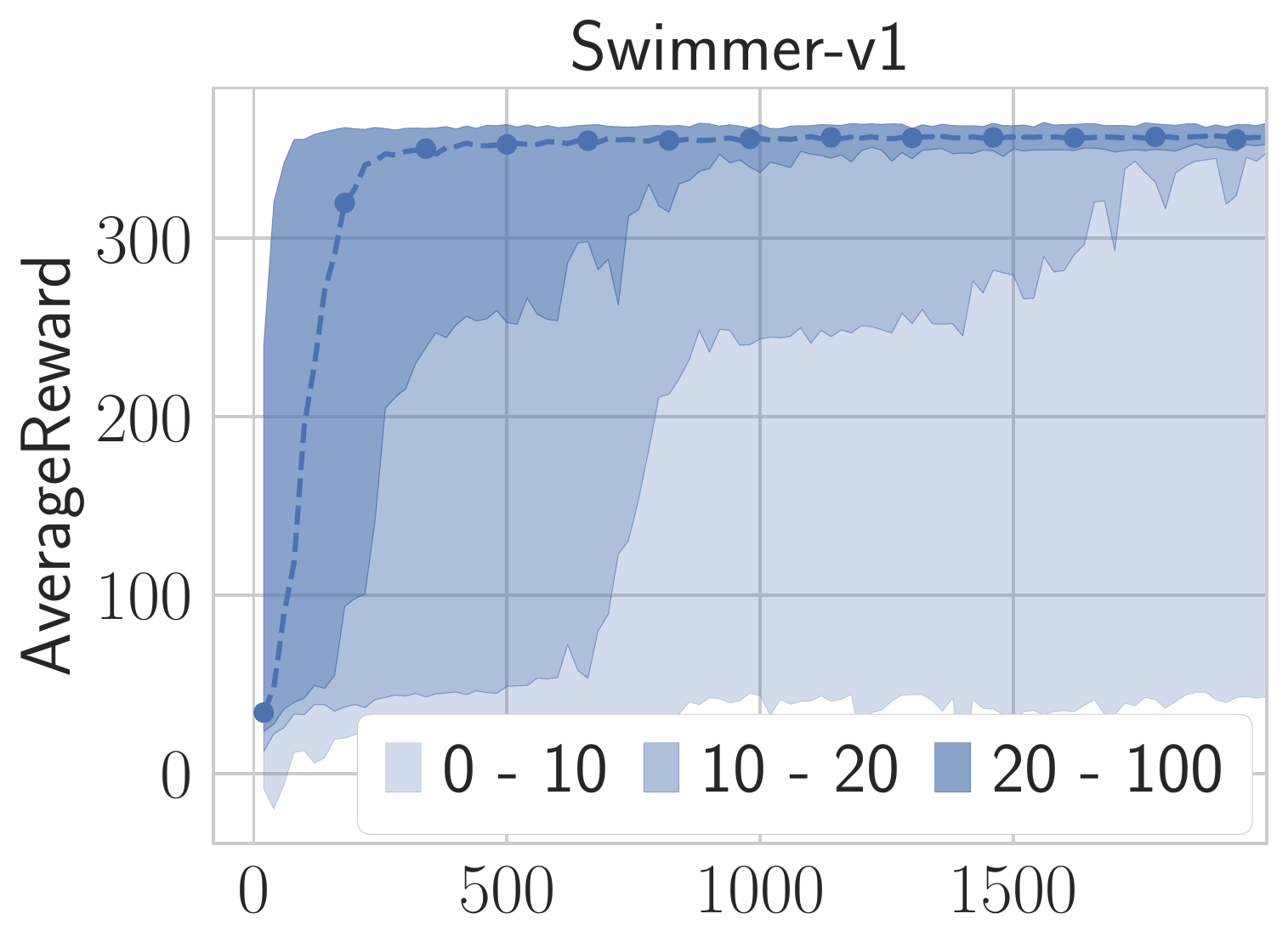}}
\end{subfigure}
\begin{subfigure}[b]{0.32\textwidth}
\centerline{\includegraphics[height=3.5cm]{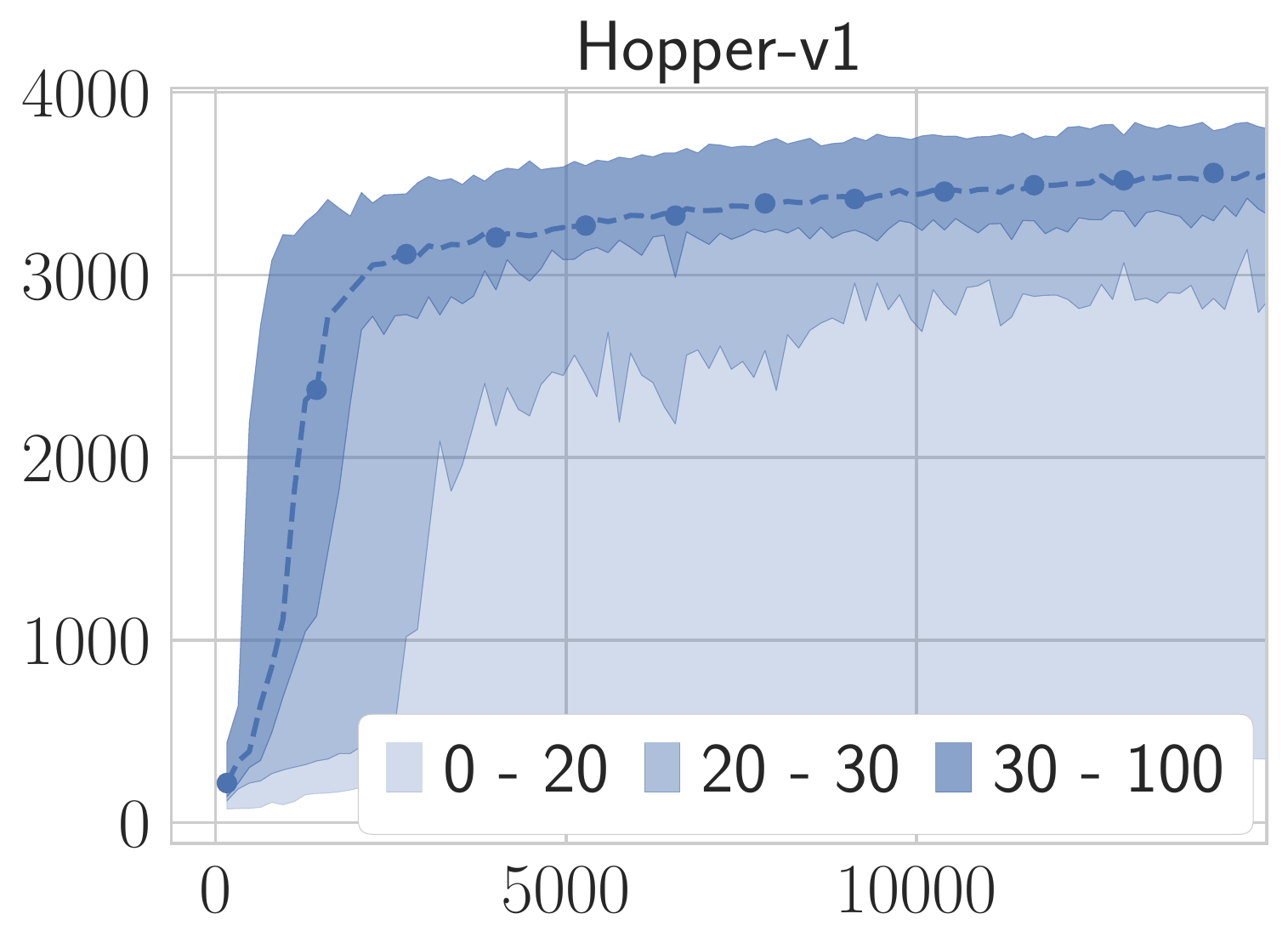}}
\end{subfigure}
\begin{subfigure}[b]{0.32\textwidth}
\centerline{\includegraphics[height=3.5cm]{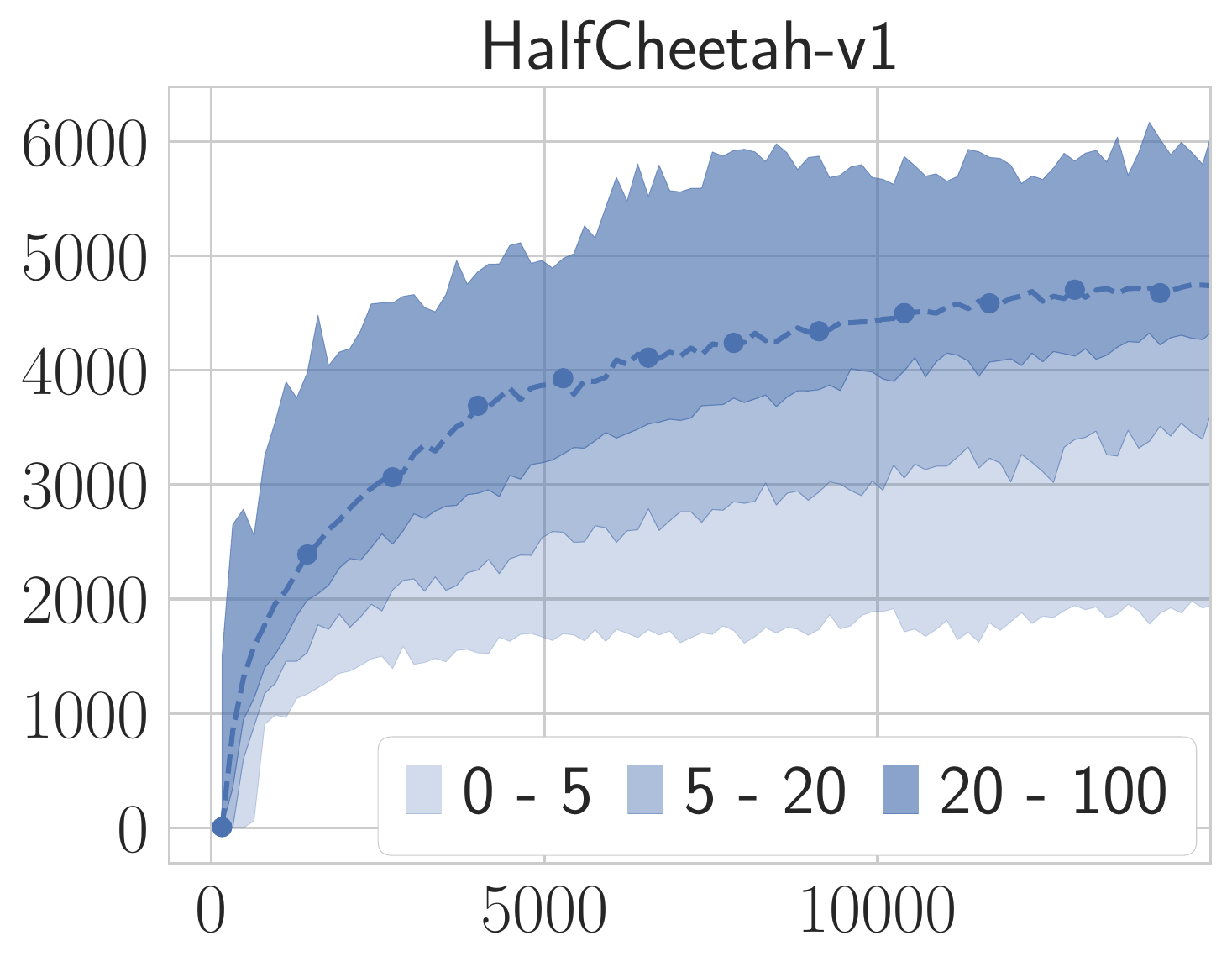}}
\end{subfigure}
\begin{subfigure}[b]{0.32\textwidth}
\centerline{\includegraphics[height=3.5cm]{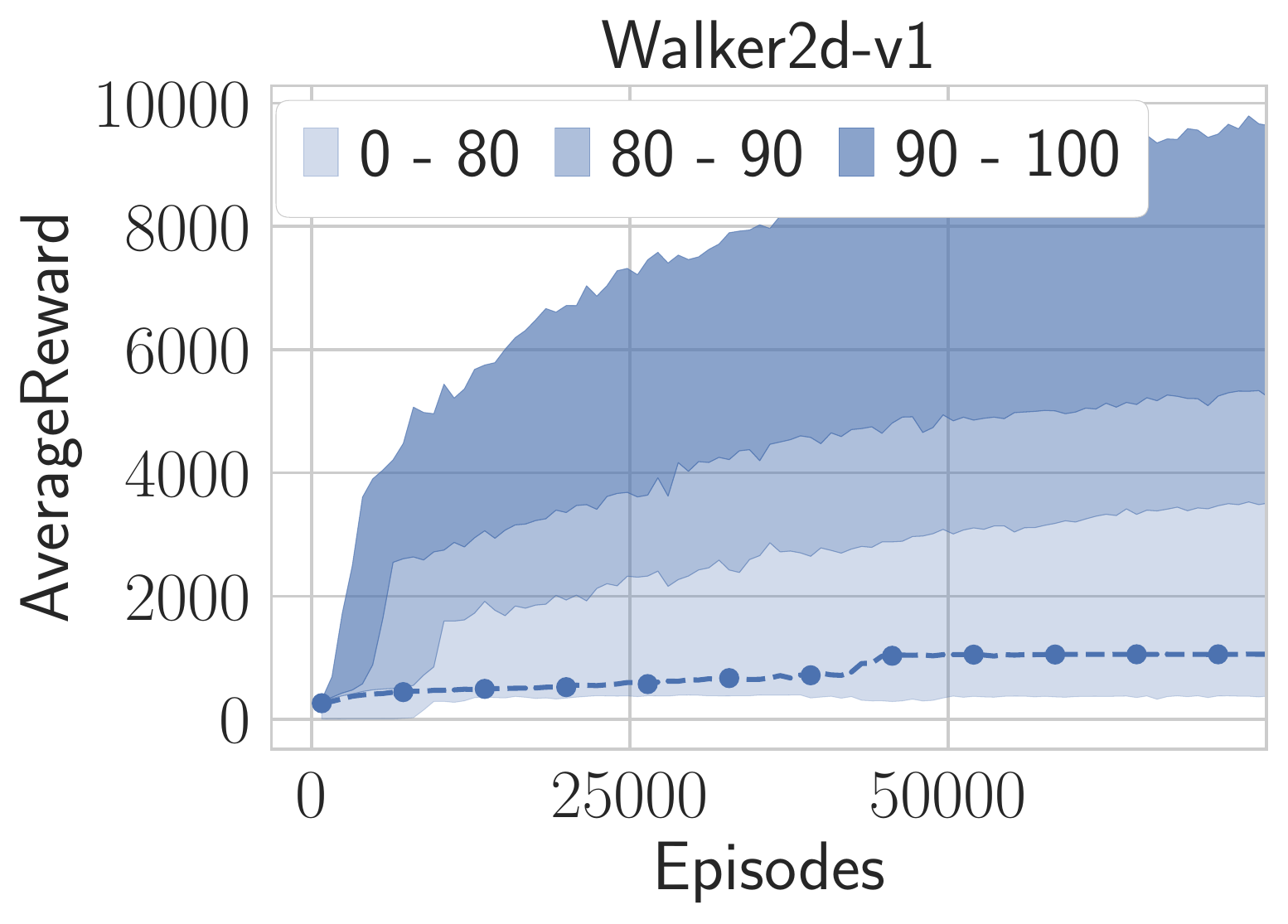}}
\end{subfigure}
\begin{subfigure}[b]{0.32\textwidth}
\centerline{\includegraphics[height=3.5cm]{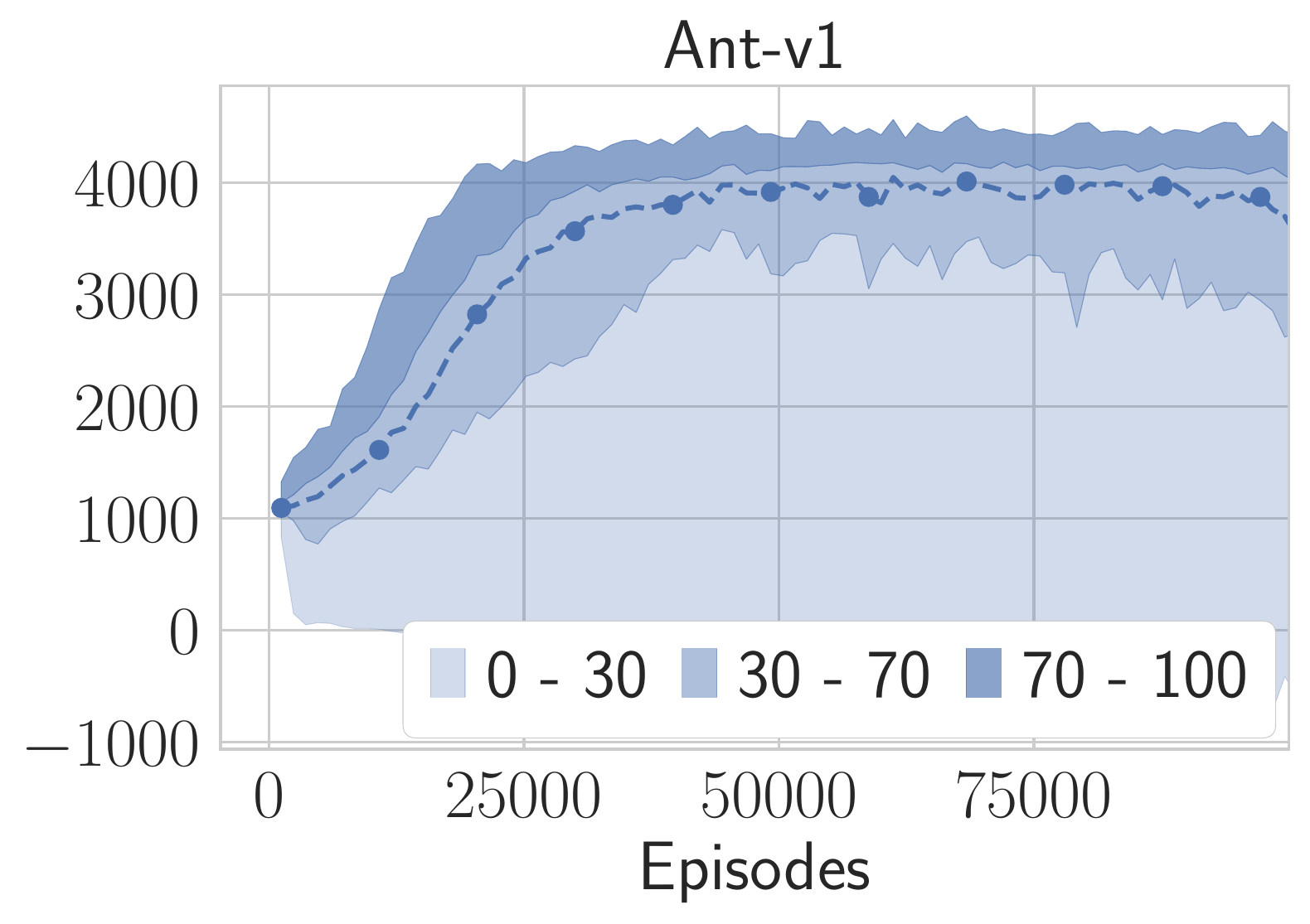}}
\end{subfigure}
\begin{subfigure}[b]{0.32\textwidth}
\centerline{\includegraphics[height=3.5cm]{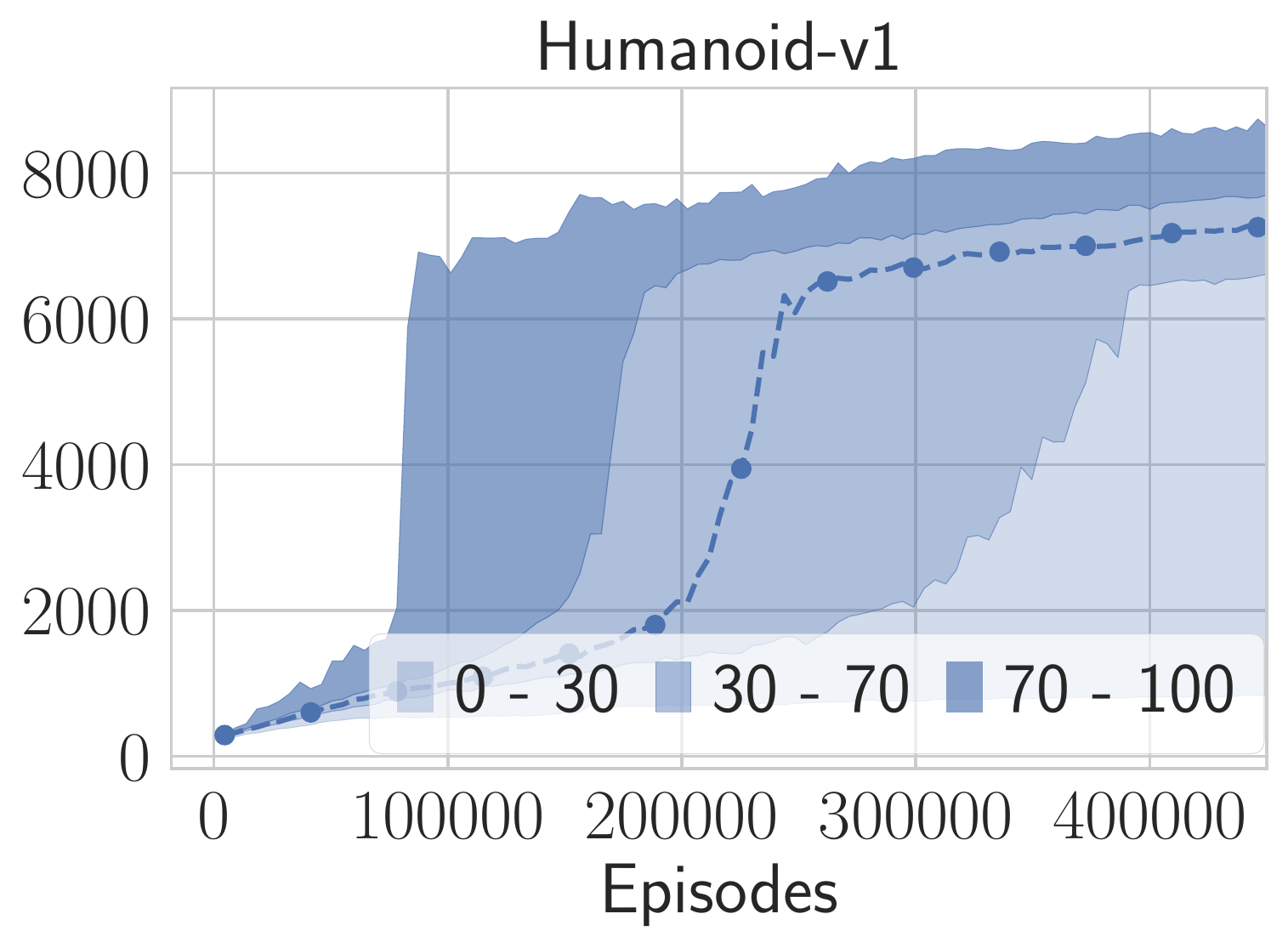}}
\end{subfigure}
\caption{An evaluation of \name{} over $100$ random seeds on the MuJoCo locomotion tasks. The dotted 
lines represent median rewards and the shaded regions represent percentiles. For Swimmer-v1 we 
used \name{} \vone{}. For Hopper-v1, Walker2d-v1, and Ant-v1 we used \name{} \vtwot{}. For HalfCheetah-v1 and Humanoid-v1 we used 
\name{} \vtwo{}.}
\label{fig:100_seeds}
\end{figure}

The results are shown in Figure~\ref{fig:100_seeds}. 
Figure~\ref{fig:100_seeds} shows that $70\%$ of the time \name{} trains policies for all the MuJoCo locomotion tasks, with the exception 
of Walker2d-v1 for which it succeeds only $20\%$ of the time. Moreover, \name{} succeeds at training policies a large fraction 
of the time while using a competitive number of episodes.

There are two types of random seeds that are used in Figure~\ref{fig:100_seeds} and that cause \name{} to not reach high rewards. 
There are random seeds on which \name{} eventually finds high reward policies if sufficiently many iterations of ARS are performed, 
and there are random seeds which lead \name{} to discover locally optimal behaviors. 
For the Humanoid model, ARS found numerous distinct gait, 
including ones during which the Humanoid hopes only in one leg, walks backwards, or moves in a swirling motion. 
Such gaits were found by ARS on the random seeds which cause slower training . 
While multiple gaits for Humanoid models have been previously observed \cite{heess2017emergence}, 
our evaluation better emphasizes their prevalence.  
These results further emphasize the importance of evaluating RL algorithms on many random seeds since evaluations
on small numbers of seeds cannot correctly capture the ability of algorithms 
to find good solutions for highly non-convex optimization problems.

Finally, Figure~\ref{fig:100_seeds} shows that \name{} is the least sensitive 
to the random seed used when applied to the HalfCheetah-v1 problem. While SAC achieved a higher  
 reward than ARS on this task, \citet{haarnoja2018soft} evaluated the sensitivity of SAC to 
random seeds only on HalfCheetah-v1.

\paragraph{Sensitivity to hyperparameters:}

\begin{figure}[h]
\centering
\caption*{\bf Evaluation of sensitivity to hyperparameters, shown by percentile}
\begin{subfigure}[b]{0.32\textwidth}
\centerline{\includegraphics[height=3.5cm]{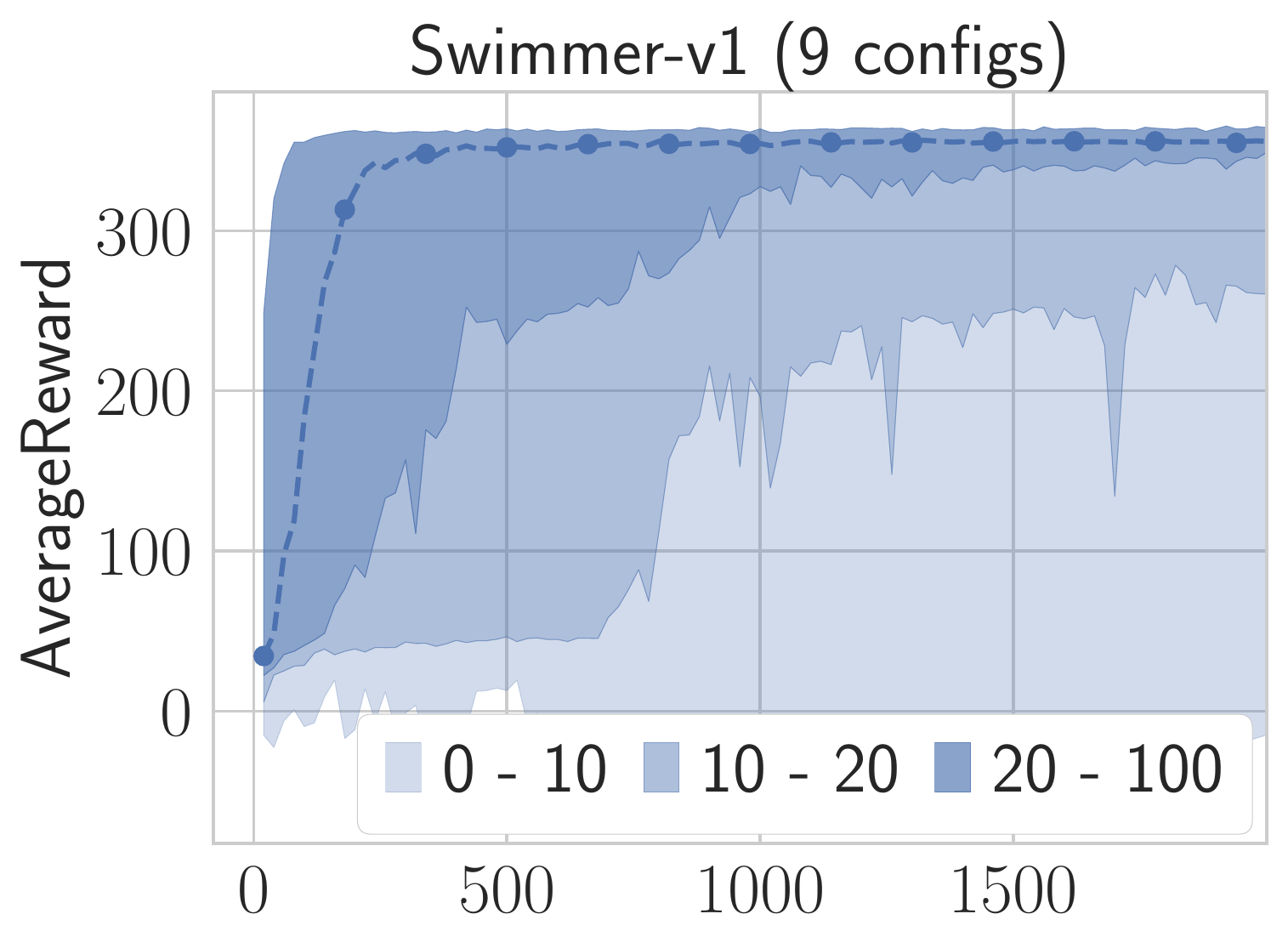}}
\end{subfigure}
\begin{subfigure}[b]{0.32\textwidth}
\centerline{\includegraphics[height=3.5cm]{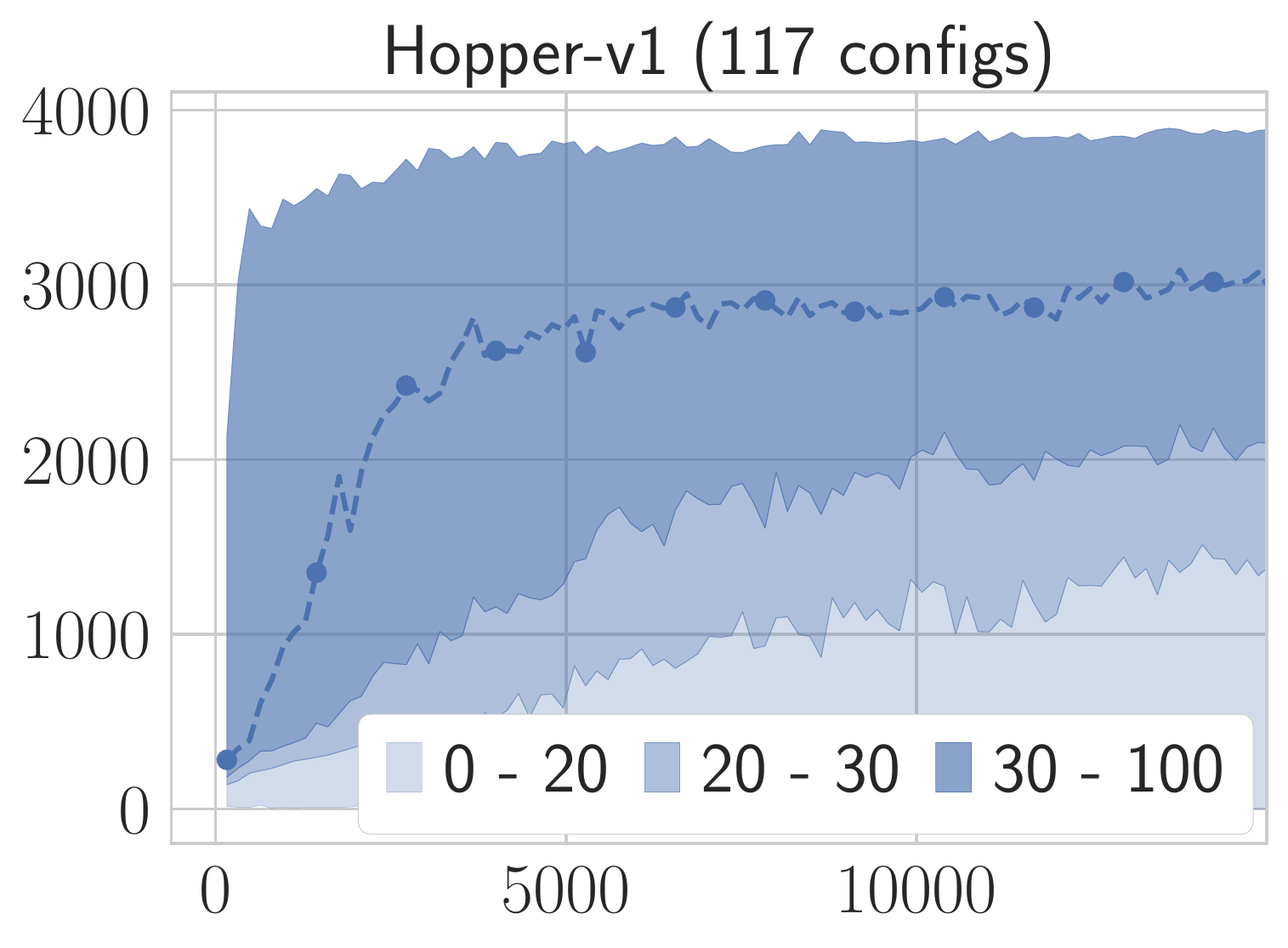}}
\end{subfigure}
\begin{subfigure}[b]{0.32\textwidth}
\centerline{\includegraphics[height=3.5cm]{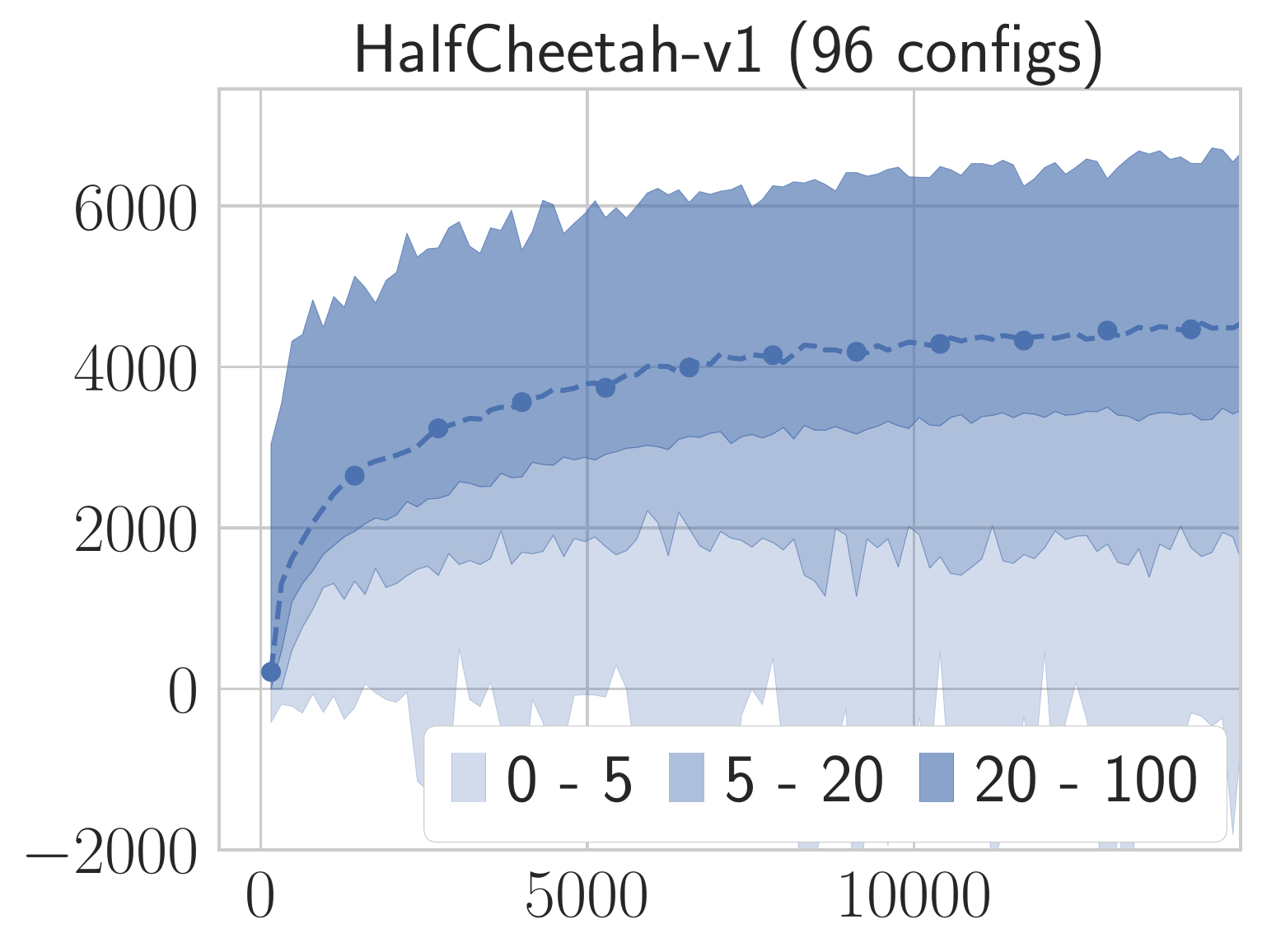}}
\end{subfigure}
\begin{subfigure}[b]{0.32\textwidth}
\centerline{\includegraphics[height=3.5cm]{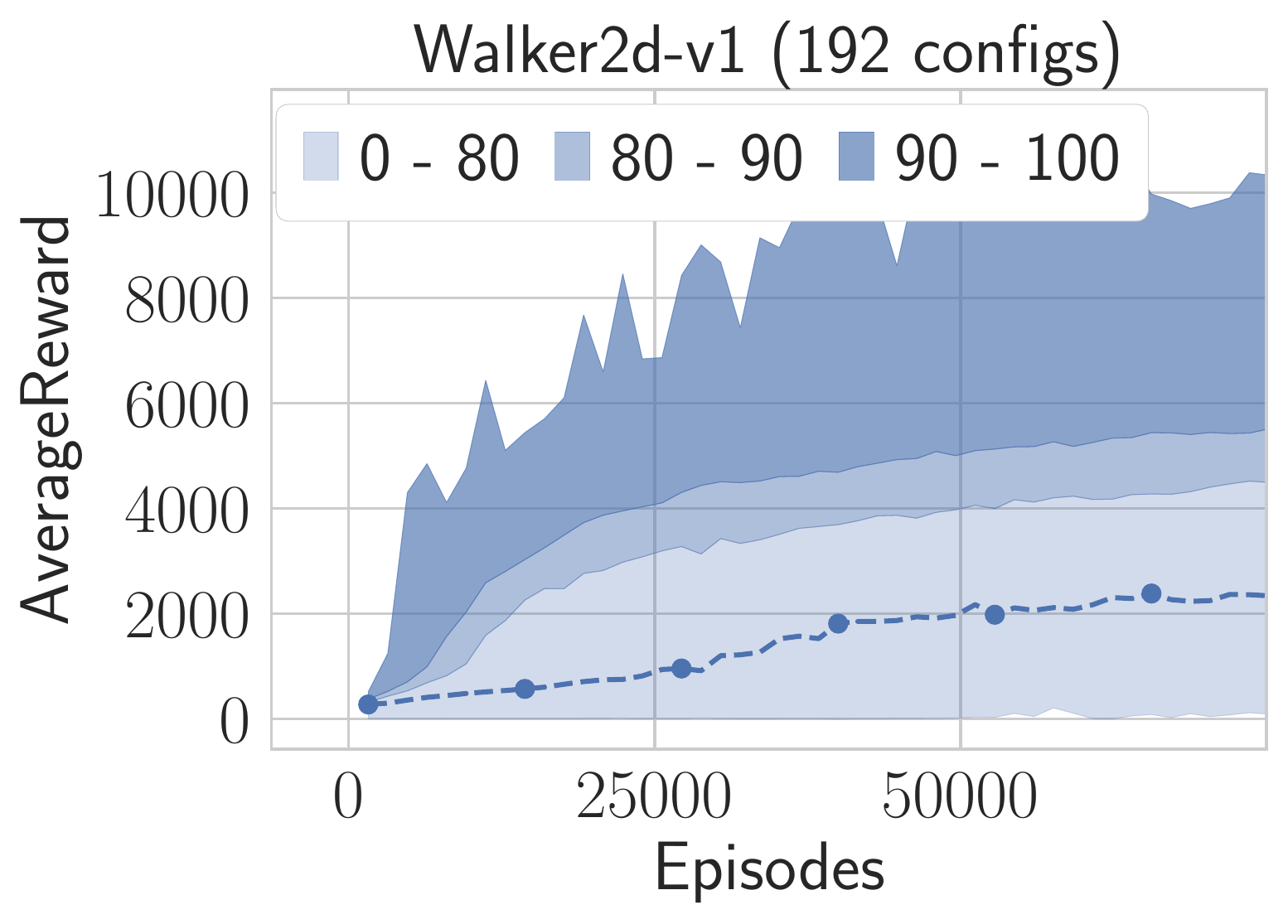}}
\end{subfigure}
\begin{subfigure}[b]{0.32\textwidth}
\centerline{\includegraphics[height=3.5cm]{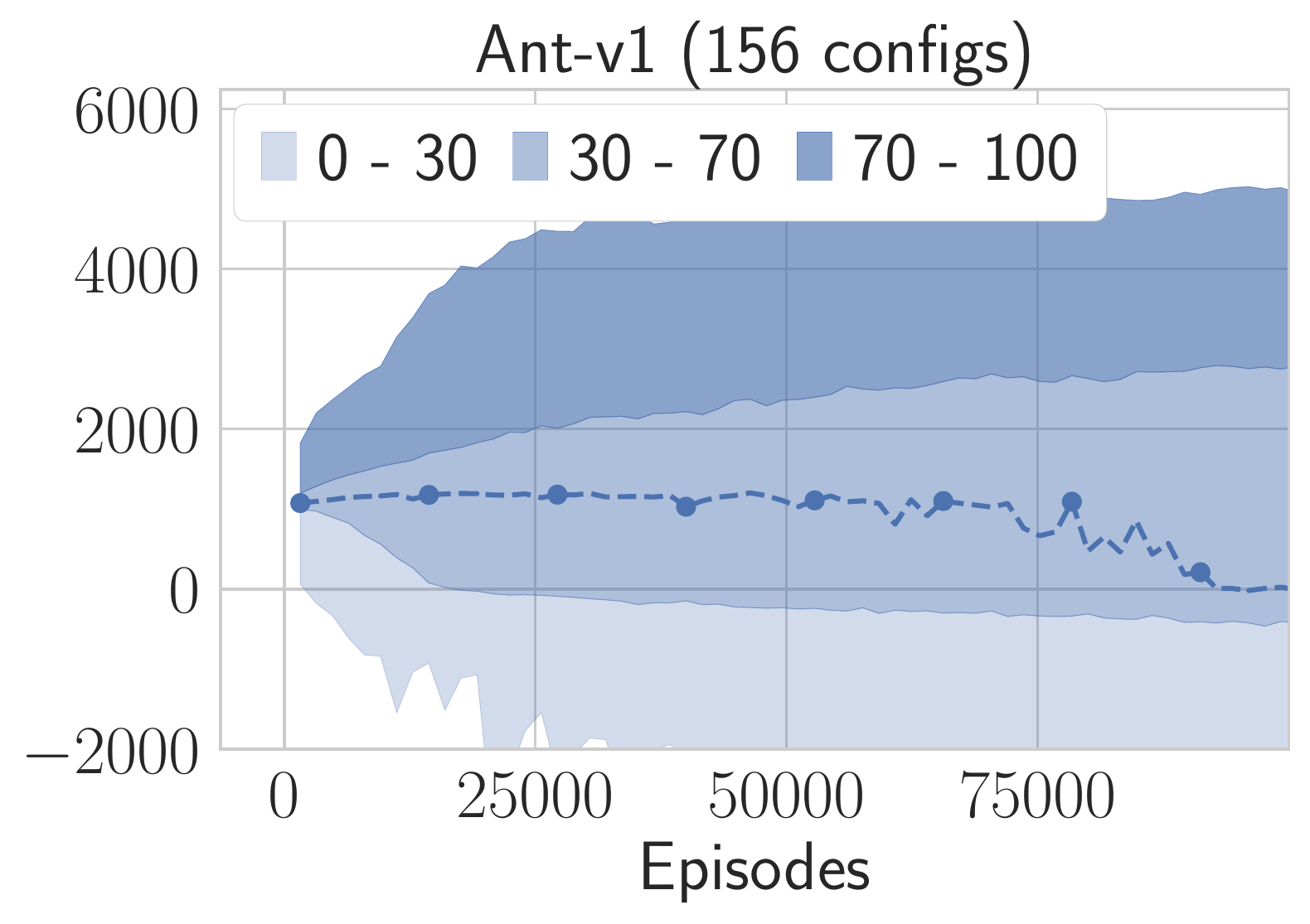}}
\end{subfigure}
\begin{subfigure}[b]{0.32\textwidth}
\centerline{\includegraphics[height=3.5cm]{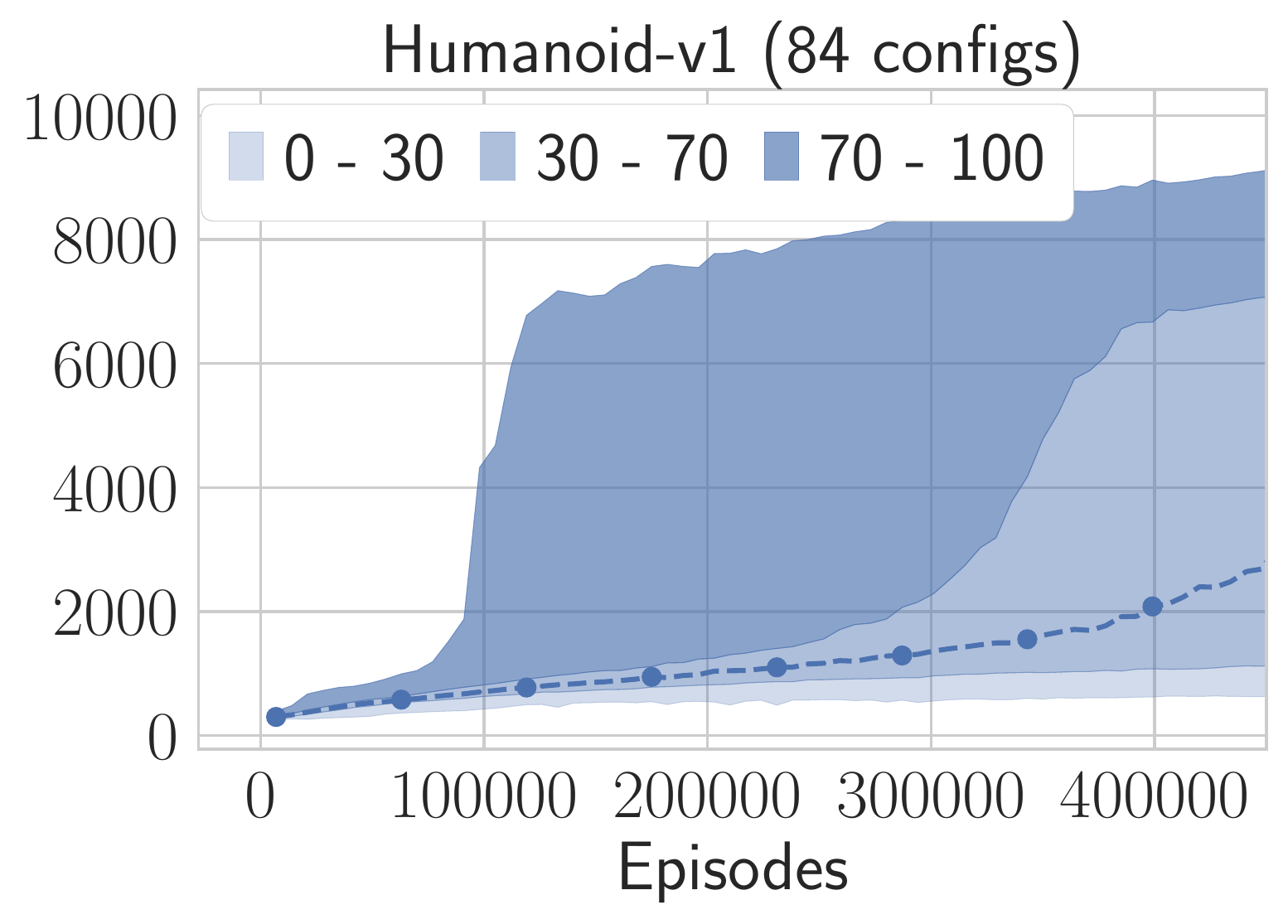}}
\end{subfigure}
\caption{An evaluation of the sensitivity of \name{} to the choice of hyperparameters. 
The dotted lines represent median average reward and the shaded regions represent percentiles.
We used all the learning curves collected during the hyperparameter tuning performed for the evaluation over the three fixed random seeds.  
For Swimmer-v1 we used \name{} \vone{}, and for the rest of the environments we used \name{} \vtwot{} (and implicitly \vtwo{} when $b = N$).}
\label{fig:parameter_perturbations}
\end{figure}

It has been correctly noted in the literature that RL methods should not be sensitive to hyperparameter choices if one hopes to apply them 
in practice \cite{haarnoja2018soft}.
For example, DDPG is known to be highly sensitive to hyperparameter choices, making it difficult to use in practice \cite{duan2016benchmarking, haarnoja2018soft, henderson2017deep}. In the evaluations of \name{} presented above we used hyperparameters chosen by tuning over the three fixed random seeds. 
To determine the sensitivity of ARS to the choice of hyperarameters, in Figure~\ref{fig:parameter_perturbations} we plot the median performance 
of all the hyperparameters considered for tuning over the three fixed random seeds. Recall that the grids of hyperparameters used
for the different MuJoCo tasks are shown in Appendix~\ref{sec:parameters}. 

Interestingly, the success rates of \name{} depicted in Figure~\ref{fig:parameter_perturbations} are similar 
to those shown in Figure~\ref{fig:100_seeds}. Figure~\ref{fig:parameter_perturbations} shows a decrease in median performance only for
Ant-v1 and Humanoid-v1. The similarity between Figures~\ref{fig:100_seeds} and \ref{fig:parameter_perturbations}
shows that the success of \name{} is as 
influenced by the choice of hyperparameters as it is by the choice of random seeds. To put it another way, 
\name{} is not highly sensitive to the choice of hyperparameters because its success rate when varying hyperarameters
is similar to its success rate when performing independent trials with a ``good'' choice of hyperparameters. 
Finally, Figure~\ref{fig:parameter_perturbations} shows that the performance of \name{} on the HalfCheetah-v1 task, a problem 
often used for evaluations of sensitivity \cite{haarnoja2018soft, henderson2017deep}, is the 
least sensitive to the choice of hyperparameter. 

\paragraph{Linear policies are sufficiently expressive for MuJoCo:} 

For our evaluation on $100$ random seeds we discussed how linear policies can produce diverse 
gaits for the MuJoCo models, showing that linear policies are sufficiently expressive to capture 
diverse behaviors. Moreover, Table~\ref{table:max_rewards} shows that linear policies 
can achieve high rewards on all the MuJoCo locomotion tasks. In particular, for the 
Humanoid-v1 and Walker2d-v1  ARS found policies that achieve significantly higher rewards
than any other results we encountered in the literature. 
These results show that linear policies are perfectly adequate for the MuJoCo locomotion tasks,
 reducing the need for more expressive and more computationally expensive policies.

\begin{table}[h]
\centering 
\begin{tabular}{cc}
 \multicolumn{2}{c}{\bf Maximum reward achieved} \\
\hline
\textbf{Task} & \textbf{\name{}} \\
\hline 
Swimmer-v1 & $365$ \\
Hopper-v1  & $3909$ \\
HalfCheetah-v1 & $6722$ \\
Walker  & $11389$ \\
Ant     & $5146$ \\
Humanoid & $11600$ \\
\hline
\end{tabular}
\caption{Maximum average reward achieved by \name{}, where we took the 
maximum over all sets of hyperparameters considered and the three fixed random seeds.}
\label{table:max_rewards}
\end{table}


\subsection{Linear quadratic regulator}
\label{sec:lqr}

While the MuJoCo locomotion tasks considered above are popular benchmarks in the RL literature, they have their shortcomings.  The maximal achievable awards are unknown as are the optimal policies, and the current state-of-the-art may indeed be very suboptimal. These methods exhibit high variance making it difficult to distinguish quality of learned policies.  And, since it is hard to generate new instances, the community may be overfitting to this small suite of tests. 

In this section we propose a simpler benchmark which obviates many of these shortcomings: the classical Linear Quadratic Regulator (LQR) with unknown dynamics. 
In control theory the LQR with known dynamics is a fundamental problem, which is thoroughly 
understood.  In this problem the goal is to control a linear dynamical system while minimizing a quadratic cost. The problem is formalized in Eq.~\eqref{eq:lqr}. 
The states $x_t$ lie in $\RR^\statedim$, the actions $u_t$ lie in $\RR^\inputdim$, and the matrices $A$, $B$, $Q$, and 
$R$ are have the appropriate dimensions. The noise process $w_t$ is i.i.d. Guassian. 
When the dynamics $(A,B)$ are known, under mild conditions, problem ~\eqref{eq:lqr} admits an optimal policy of the form $u_t = Kx_t$ for some unique matrix $K$, computed efficiently from the solution of an algebraic Riccati equation.
Moreover, the finite horizon version of problem~\ref{eq:lqr} can be efficiently solved via dynamic programming.

\begin{align}
\label{eq:lqr}
&\min_{u_0, u_1, \ldots}\: \lim_{T \rightarrow \infty} \frac{1}{T}\EE \left[\sum_{t = 0}^{T - 1} x_t^\top Q x_t + u_t^\top R u_t\right]\\
&\text{s.t. } \; x_{t + 1} = A x_t + B u_t + w_t
\nonumber
\end{align}

LQR with unknown dynamics is considerably less well understood and offers a fertile ground for new research. Note that it is still trivial to produce a varied set of instances for LQR, and we can always compare the best achievable cost when the dynamics are known. 

A natural model-based approach consists of estimating the transition matrices $(A,B)$ from data 
and then solving for $K$ by plugging the estimates in the Riccati equation. 
A controller $K$ computed in this fashion is called a \emph{nominal controller}. Though this method may not be ideally robust (see, e.g., \citet{dean2017sample}), nominal control provide a useful baseline to which we can compare other methods.

\begin{figure} 
\centering
\begin{subfigure}[t]{0.49\textwidth}
\centerline{\includegraphics[width=1.\columnwidth]{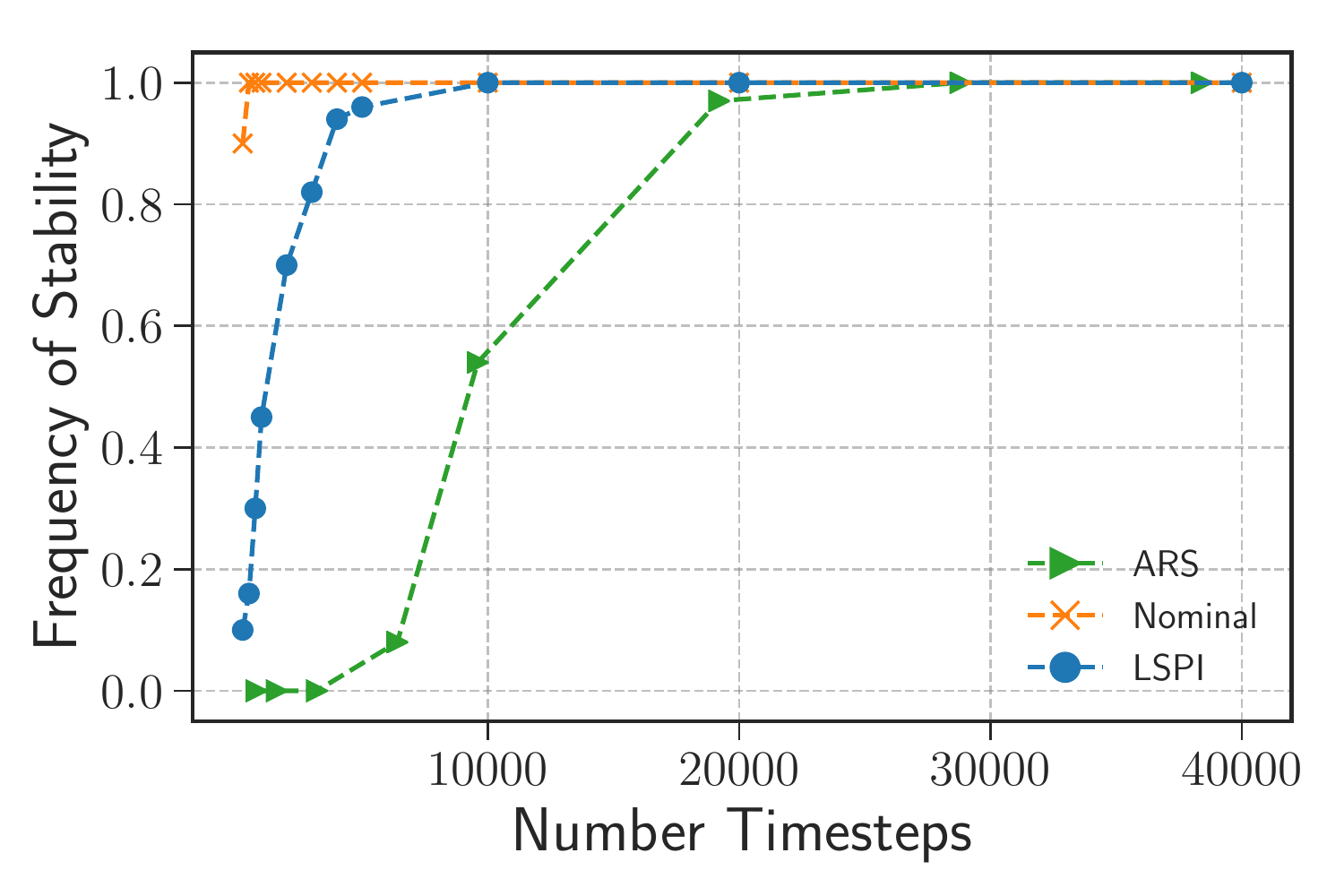}}
\caption{A comparison of how frequently the controllers produced by ARS, the nominal
synthesis procedure, and the LSPI method find stabilizing controllers. 
The frequencies are estimated from $100$ trials.}
\label{fig:stabilizing}
\end{subfigure}
\begin{subfigure}[t]{0.49\textwidth}
\centerline{\includegraphics[width=1.\columnwidth]{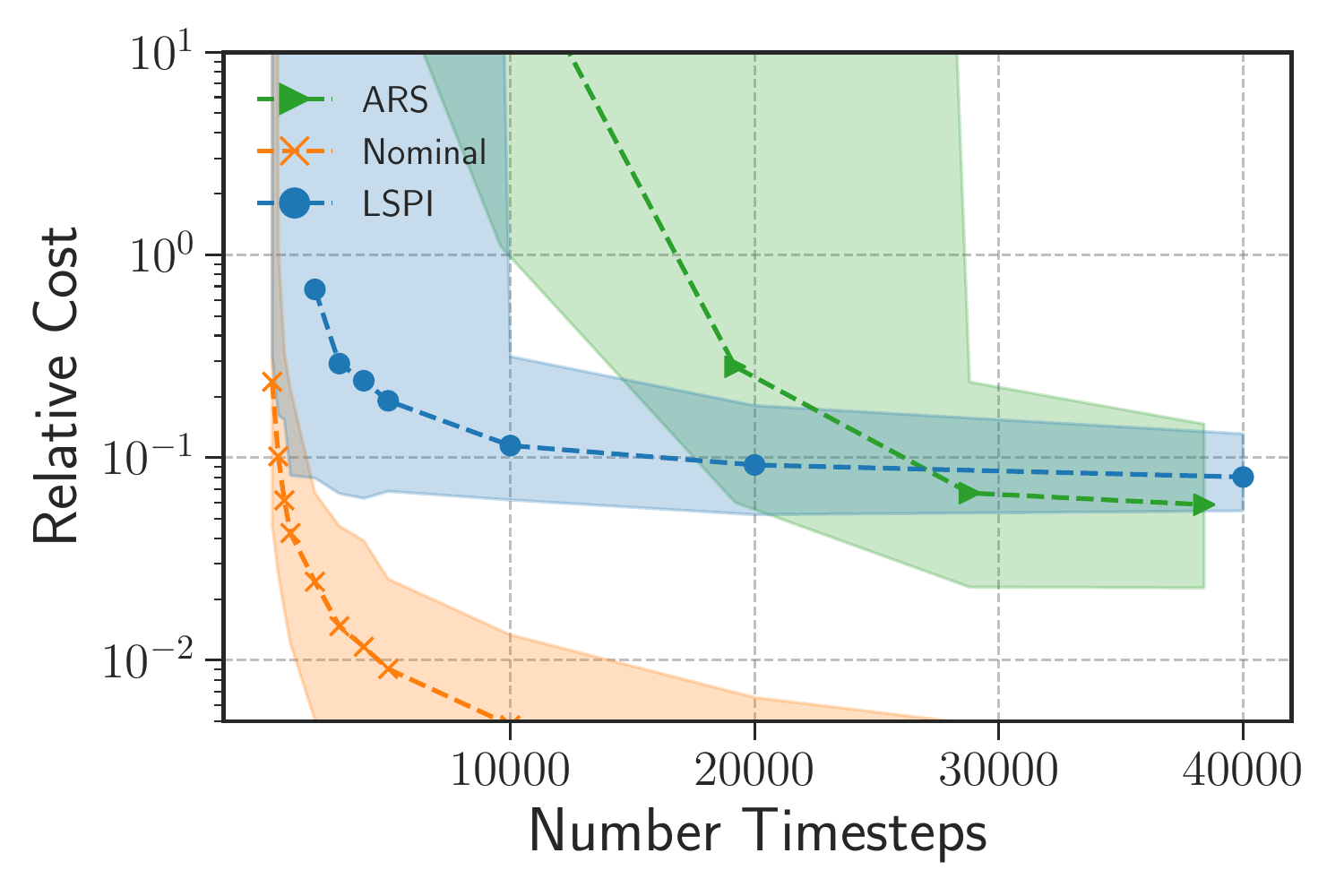}}
\caption{A comparison of the relative cost of the controllers produced by ARS, the nominal
synthesis procedure, and the LSPI method. 
The points along the dashed line denote the
median cost, and the shaded region covers the
$2$-nd to $98$-th percentile out of $100$ trials.}
\label{fig:rel_cost}
\end{subfigure}
\caption{A comparison of four methods when applied to the LQR problem~\eqref{eq:exampledynamics}.}
\label{fig:lqr}
\end{figure}

Consider the LQR instance defined introduced by \citet{dean2017sample} as a challenging low-dimensional instance for LQR with unknown dynamics. 
\begin{align} \label{eq:exampledynamics}
A  = \begin{bmatrix} 1.01 & 0.01 & 0\\
0.01 & 1.01 & 0.01\\
0 & 0.01 & 1.01\end{bmatrix}, ~~ B = I, ~~ Q = 10^{-3} I, ~~ R =I \:.
\end{align}
The matrix $A$ has eigenvalues greater than $1$, and hence the system is unstable without some control. Moreover, if a method fails to recognize that the system is unstable, it may not yield a stable controller. 
In Figure~\ref{fig:lqr} we compare \name{} to nominal control and to a method using Q-functions fitted by temporal differencing (LSPI), analyzed by \citet{tu2017least}.

While Figure~\ref{fig:rel_cost} shows \name{} to require significantly more samples than LSPI to find a stabilizing controller, we note that LSPI requires an initial controller $K_0$ which stabilizes a discounted version of problem~\eqref{eq:lqr}. \name{} does not require a special initialization. 
However, Figure~\ref{fig:rel_cost} also shows that the nominal control method is orders of magnitude more sample efficient than both LSPI and \name{}. Hence there is much room for improvement for pure model-free approaches.

We conjecture that the LQR instance~\eqref{eq:exampledynamics} would also be particularly challenging for policy gradient methods or other methods that explore in the action space. When the control signal is zero, the linear system described by Eq.~\eqref{eq:exampledynamics} has a small spectral radius ($\rho \approx 1.024$) and as a result the states $x_t$ would blow up, but slowly. Therefore long trajectories are required for evaluating the performance of a controller. However, the variance of policy gradient methods grows with the length of the trajectories used, even when standard variance reduction techniques are used. 
 
\subsection{Computational efficiency}
\label{sec:computation}

The small computational footprint of linear policies and the embarrassingly parallel structure of \name{} make our method ideal 
for training policies in a small amount of time, with few computational resources. In Tables~\ref{table:es_time} and \ref{table:percentile_time}
we show the wall-clock time required by \name{} to reach an average reward of $6000$, evaluated over $100$ random seeds. 
\name{} requires a median time of $21$ minutes to reach the prescribed reward threshold, when trained on one m5.24xlarge EC2 instance with $48$ CPUs. 
The Evolution Strategies method of \citet{salimans2017evolution} took a median time of $10$ minutes when evaluated over $7$ trials. However, the authors do not clarify what $7$ trails means, multiple trials with the same random seed or multiple trials with different random seeds. Moreover, 
Table~\ref{table:percentile_time} shows that \name{} trains a policy in at most $10$ minutes on $10$ out of $100$ seeds. 
Also, Table~\ref{table:es_time} shows that \name{} requires up to $15$ times less CPU time than ES. 

Finally, we would like to point out that our method could be scaled to more workers. In that case, \name{} \vtwo{} will have a computational bottleneck 
 in the aggregation of the statistics $\Sigma_j$ and $\mu_j$ across workers. 
For successful training of policies, \name{} \vtwo{} does not require the update of the statistics (see Line~\ref{line:update_stats} of Algorithm~\ref{alg:lin-rl}) to occur at each iteration. 
For example, in their implementation of ES, \citet{moritz2017ray} allowed each worker to have its own independent estimate of $\mu_j$ and $\Sigma_j$. With this choice, the authors used Ray to scale ES to $8192$ cores, reaching a $6000$ reward 
 on Humanoid-v1 in $3.7$ minutes. 
One could tune an update schedule for the statistics $\Sigma_j$ and $\mu_j$ in order to reduce the communication 
time between workers or reduce the sample complexity of ARS. For the sake of simplicity we refrained from tuning such a schedule. 

\begin{table}[H]
\centering 
\begin{tabular}{ccccc}
\hline
\textbf{Algorithm} & \textbf{Instance Type} & \textbf{\# CPUs} & \textbf{Median Time} & \textbf{CPU Time}  \\
\hline 

\multirow{2}{*}{Evolution Strategies} & UNK & $18$ & $657$ minutes & $197$ hours \\
 & UNK & $1440$ & $10$ minutes & $240$ hours \\
\hline 
\multirow{3}{*}{ARS \vtwo{}} & m5.24xlarge  & $48$ &  $21$ minutes & $16$ hours \\
 & c5.9xlarge  & $18$ & $41$ minutes & $12$ hours \\
 & c4.8xlarge & $18$ & $57$ minutes & $17$ hours \\
\hline
\end{tabular}
\caption{An evaluation of the wall-clock time required to reach an average reward of $6000$ for the Humanoid-v1 task. The median time required by \name{} was evaluated over $100$ random seeds. 
The values for ES were taken from the work by \citet{salimans2017evolution}, and were evaluated over $7$ independent trials. UNK stands for unknown.}
\label{table:es_time}
\end{table}

\begin{table}[H]
\centering 
\begin{tabular}{ccc|cccc}
& & & \multicolumn{4}{c}{{\bf \# minutes by percentile}}\\
\hline
\textbf{Algorithm} & \textbf{Instance Type} & \textbf{\# CPUs} & \textbf{10th} & \textbf{25th} & \textbf{50th} & \textbf{75th}\\
\hline 
\multirow{3}{*}{ARS \vtwo{}} & m5.24xlarge & $48$ & $10$ & $13$  & $21$  & $45$ \\
                   & c5.9xlarge  & $18$ & $16$ & $23$ & $41$ & $96$ \\
                   & c4.8xlarge  & $18$ & $21$ & $28$ & $57$ & $144$ \\
\hline
\end{tabular}
\caption{A breakdown by percentile of the number of minutes required by \name{} to reach an average reward of $6000$ on the Humanoid-v1 task. 
The percentiles were computed over runs on $100$ random seeds.}
\label{table:percentile_time}
\end{table}


\section{Conclusion}\label{sec:discussion}

We attempted to find the simplest algorithm for model-free RL that performs well on the continuous control benchmarks used in the RL literature. We demonstrated that with a few algorithmic augmentations, basic random search could be used to train \emph{linear} policies that achieve state-of-the-art sample efficiency on the MuJoCo locomotion tasks. We showed that linear policies match the performance of complex neural network policies and can be found through a simple algorithm. Since the algorithm and policies are simple, we were able to perform extensive sensitivity studies, and observed that our method can find good solutions to highly nonconvex problems a large fraction of the time. Up to the variance of RL algorithms \cite{henderson2017deep, islam2017reproducibility}, our method achieves state-of-the-art performance on the MuJoCo locomotion tasks when hyperparameters and random seeds are varied. 
Our results emphasize the high variance intrinsic to the training of policies for MuJoCo RL tasks. Therefore, it is not clear what is gained by evaluating RL algorithms on only a small numbers of random seeds, as is common in the RL literature. Evaluation on small numbers of random seeds 
does not capture performance adequately due to high variance.

Our results point out some problems with the common methodology used for the evaluation of RL algorithms.  Though many RL researchers are concerned about minimizing sample complexity, \emph{it does not make sense to optimize the running time of an algorithm on a single instance.} The running time of an algorithm is only a meaningful notion if either (a) evaluated on a family of instances, or (b) when clearly restricting the class of algorithms. 

Common RL practice, however, does not follow either (a) or (b). Instead researchers run algorithm $\Acal$
on task $\Tcal$ with a given hyperparameter configuration, and plot a ``learning curve'' showing the algorithm reaches a target reward after collecting $X$ samples. Then the ``sample complexity" of the method is reported as the number of samples required to reach a target reward threshold, with the given hyperparameter configuration.  However, any number of hyperparameter configurations can be tried. Any number of algorithmic enhancements can be added or discarded and then tested in simulation.  For a fair measurement of sample complexity, should we not count the number of rollouts used for every tested hyperparameters? 

Let us look what would happen if the field of convex optimization relied exclusively on the same method of evaluation. Suppose we wanted to assess the performance of the stochastic gradient method at optimizing the objective $G(x) = \EE g(x, \xi)$, where $g(\cdot, \xi)$ is a one dimensional function, strongly convex and smooth for all random variables $\xi$.  Moreover, let us assume that for all $\xi$ the functions $g(\cdot, \xi)$ have the same minimizer. At each iteration the algorithm queries an oracle for a stochastic gradient. The oracle samples $\xi$ and returns to the algorithm the derivative $g'(x, \xi)$. 


Then, according to the current RL methodology for evaluation, we fix a sequence of random variables
$\xi$ sampled by the oracle by fixing a random seed and then proceed to tune the step-size of the stochastic gradient method.
Since for each step-size the first random variable sampled by the oracle is the same, call it $\xi_0$,
after $\Ocal(1/\epsilon)$ tries we can determine a step-size which ensures reaching an $\epsilon$-close minimizer of 
$g(\cdot, \xi)$ after one iteration of the algorithm. 
Since all functions $g(\cdot, \xi)$ have the same minimizer, the point found after one iteration would be $\epsilon$-close to the minimizer of $G$. 
Hence, by using $\Ocal(1/\epsilon)$ oracle calls behind the scene, we can elicit a step-size which optimizes the objective in one iteration.
However, we would not say that the sample complexity of the algorithm is one because the same algorithm would require more than one sample
to optimize new objectives. 
A better measure of sample complexity would be the total number of samples required for tuning and the final optimization, which is $\Ocal(1/\epsilon)$,
but this methodology would also not capture the correct sample complexity of the algorithm. 
Indeed, we know the sample complexity of the stochastic gradient method for the optimization of objectives like $G$ is  $\Ocal(\log(1/\epsilon))$.

RL tasks are not as simple as the one dimensional convex objective considered above, and RL methods are evaluated on more than one random seed.  However, our arguments are just as relevant. 
Through optimal hyperparameter tuning one can artificially improve the perceived sample efficiency of a method.  Indeed, this is what we see in our work. By adding a third algorithmic enhancement to basic random search (i.e., enhancing ARS \vtwo{} to \vtwot{}), we are able to improve the sample efficiency of an already highly performing method.  Considering that most of the prior work in RL uses algorithms with far more tunable parameters and neural nets whose architectures themselves are hyperparameters, the significance of the reported sample complexities for those methods is not clear.  This issue is important because a meaningful sample complexity of an algorithm should inform us on the number of samples required to solve a new, previously unseen task.  A simulation task should be thought of as an \emph{instance} of a problem, not the problem itself. 


In light of these issues and of our empirical results, we make several suggestions for future work:
\begin{itemize}
\item Simple baselines should be established before moving forward to more complex benchmarks and methods. 
Simpler algorithms are easier to evaluate empirically and understand theoretically. We propose that LQR is a reasonable baseline as this task is very well-understood when the model is known, instances can be generated with a variety of different levels of difficulty, and little overhead is required for replication.

\item When games and physics simulators are used for evaluation, separate problem instances should be used for tuning and evaluation of RL methods.
Moreover, large numbers of random seeds should be used for statistically significant evaluations. 
However, since the distribution of problems occurring in games and physics simulators differs from the distribution of problems one hopes to solve, 
this methodology is not ideal either.  In particular, it is difficult to say that one algorithm is better than another when evaluations are performed only in simulation since one of the algorithms might be exploiting particularities of the simulator used. 
 
\item Rather than trying to develop algorithms which are applicable to many different classes of problems, it might be better to focus on specific problems of interest and find targeted solutions. 

\item More emphasis should be put on the development of model-based methods. For many problems, such methods have been observed to require fewer samples 
than model-free methods. Moreover, the physics of the systems should inform the parametric classes of models used for different problems. Model-based methods incur many computational challenges themselves, and it is quite possible that tools from deep RL such as improved tree search can provide new paths forward for  tasks that require the navigation of complex and uncertain environments.

\end{itemize}

\subsection*{Acknowledgments}
 
We thank Orianna DeMasi, Moritz Hardt, Eric Jonas, Robert Nishihara,  Rebecca Roelofs,  Esther Rolf, Vaishaal Shankar, Ludwig Schmidt, Nilesh Tripuraneni, Stephen Tu for many helpful comments and suggestions.
HM thanks Robert Nishihara and Vaishaal Shankar for sharing their expertise in parallel computing. 
As part of the RISE lab, HM is generally supported in part by NSF CISE Expeditions
Award CCF-1730628, DHS Award HSHQDC-16-3-00083, and gifts from Alibaba, Amazon Web
Services, Ant Financial, CapitalOne, Ericsson, GE, Google,
Huawei,  Intel,  IBM, Microsoft,  Scotiabank,  Splunk and
VMware. 
BR is generously supported in part by NSF award CCF-1359814, ONR awards N00014-14-1-0024 and N00014-17-1-2191, the DARPA Fundamental Limits of Learning (Fun LoL) Program, and an Amazon AWS AI Research Award.

\begin{small}
\bibliographystyle{abbrvnat}
\bibliography{dfo}
\end{small}

\appendix

\section{Appendix}

\subsection{Maximum reward achieved after a prescribed number of timesteps}
\label{sec:timestep_comparison}

We explain our procedure for obtaining the maximum reward achieved by ARS after a
prescribed number of timesteps, averaged over three random seeds. A natural method consists of finding the maximum reward
achieved by ARS on each random seed and averaging those values. However, 
in Tables~\ref{table:ppo} and \ref{table:sac} we compare the performance of ARS 
to results taken from the figures presented by \citet{schulman2017proximal} and \citet{haarnoja2018soft},
who average training curves across random seeds. 
For a fair comparison we cannot compare the average of maxima with the maximum of an average of training curves. Therefore, we use the following method for estimating the average maximum reward achieved by ARS. 
  
We begin by introducing some notation. Let $R_i^{(j)}$ be the reward achieved by ARS at iteration $i$, on the $j$th random seed. Also let $h_i^{(j)}$ be the total number of timesteps sampled by ARS up to iteration $i$, on the $j$th random seed. Then, we average the training curves of ARS across the three random seeds to obtain
\begin{align*}
\overline{R}_i = \frac{R_i^{(1)} + R_i^{(2)} + R_i^{(3)}}{3}. 
\end{align*}

If $\beta$ is the prescribed budget of timesteps, let $\overline{h} = \min \{i | \max\{h_i^{(1)}, h_i^{(1)}, h_i^{(1)}\} \geq \beta \}$. Then, in Tables~\ref{table:ppo} and \ref{table:sac} we report 
\begin{align}
\overline{R}_{\max} = \max_{0 \leq i\leq \overline{h}} \overline{R}_i. 
\label{eq:r_max_avg}
\end{align}

The estimate \eqref{eq:r_max_avg} is a conservative measure of the performance of ARS because $\overline{h}$ is the minimum over the random seeds of the number of iterations needed by ARS to deplete the available budget of timesteps. 

\newpage 

\subsection{Hyperparameters}
\label{sec:parameters}

\begin{table}[h!]
\centering 
\begin{tabular}{rl|rl|rl}
\hline
\multicolumn{2}{c}{Swimmer-v1} & \multicolumn{2}{c}{Hopper-v1} & \multicolumn{2}{c}{HalfCheetah-v1} \\
\hline
$\alpha:$ & 0.01, 0.02, 0.025 & $\alpha:$ & 0.01, 0.02, 0.025 & $\alpha:$ & 0.01, 0.02, 0.025 \\
$\nu:$ & 0.03, 0.02, 0.01 & $\nu:$ & 0.03, 0.025, 0.02, 0.01 & $\nu:$ & 0.025, 0.02, 0.01 \\
$N:$ & 1 & $N:$ & 8, 16, 32 & $N:$ & 4, 8, 16, 32  \\
$b:$ & 1 & $b:$ & 4, 8, 32 & $b:$ & 2, 4, 8, 32 \\
\hline
\multicolumn{2}{c}{Walker-v1} & \multicolumn{2}{c}{Ant-v1} & \multicolumn{2}{c}{Humanoid-v1} \\
\hline
$\alpha:$ & 0.01, 0.02, 0.025, 0.03 &  $\alpha:$ & 0.01, 0.015, 0.02, 0.025 & $\alpha:$ & 0.01, 0.02, 0.025 \\
$\nu:$ & 0.025, 0.02, 0.01, 0.0075 & $\nu:$ & 0.025, 0.02, 0.01 & $\nu:$ & 0.01, 0.0075 \\
$N:$ & 40, 60, 80, 100 &  $N:$ & 20, 40, 60, 80 & $N:$ & 90, 230, 270, 310, 350 \\
$b:$ & 15, 30, 100 & $b:$ & 15, 20, 40, 80 & $b:$ & 100, 200, 360 \\
\hline
\end{tabular}
\caption{{ Grids of hyperparameters used during hyperarameter tuning.}}
\end{table}

\begin{table}[h!]
	\begin{tabular}{cccc ||| ccccc}
		& \vtwo{} &&& \multicolumn{5}{c}{\vtwot{}} \\
		\hline
		\textbf{Task} & $\alpha$ & $\nu$ & $N$ & \textbf{Task} & $\alpha$ & $\nu$ & $N$ & $b$ \\
		\hline
		Swimmer-v1 & 0.02 & 0.01 & 1 & Swimmer-v1 & 0.02 & 0.01 & 1 & 1 \\
		Hopper-v1 & 0.02 & 0.02 & 4 & Hopper-v1 & 0.01 & 0.025 & 8 & 4 \\
		HalfCheetah-v1 & 0.02 & 0.03 & 8 & HalfCheetah-v1 & 0.02 & 0.03 & 32 & 4 \\
		Walker2d-v1 & 0.025 & 0.01 & 60 & Walker2d-v1 & 0.03 & 0.025 & 40 & 30 \\
		Ant-v1 & 0.01 & 0.025 & 40 & Ant-v1 & 0.015 & 0.025 & 60 & 20 \\
		Humanoid-v1 & 0.02 & 0.0075 & 230 & Humanoid-v1 & 0.02 & 0.0075 & 230 & 230\\
		\hline
	\end{tabular}
	\caption{{Hyperparameters for ARS \vtwo{} and \vtwot{} used for the results shown Figure~\ref{fig:3_seeds}.}}
\end{table}


\end{document}